\documentclass{article}

    \usepackage[preprint]{neurips_2024}

\usepackage[utf8]{inputenc} 
\usepackage[T1]{fontenc}    
\usepackage{hyperref}       
\usepackage{url}            
\usepackage{booktabs}       
\usepackage{amsfonts}       
\usepackage{nicefrac}       
\usepackage{microtype}      
\usepackage{xcolor}         

\usepackage{amsmath}
\usepackage{amssymb}
\usepackage{mathtools}
\usepackage{amsthm}

\usepackage{times}
\usepackage{epsfig}
\usepackage{graphicx}
\usepackage{amsmath}
\usepackage{amssymb}

\usepackage{multirow}
\usepackage{booktabs}
\usepackage{subfloat}
\usepackage{subfig}
\usepackage{xcolor}
\usepackage{enumitem}
\usepackage{algorithm}
\usepackage{float}
\usepackage{wrapfig}

\usepackage{algpseudocode}
\renewcommand{\algorithmicrequire}{\textbf{Input:}}
\renewcommand{\algorithmicensure}{\textbf{Output:}}

\theoremstyle{plain}

\theoremstyle{definition}

\theoremstyle{remark}

 \title{CoopHash: Cooperative Learning of Multipurpose Descriptor \& Contrastive Pair Generator via Variational MCMC Teaching for Image Hashing} 

\author{
\textbf{Khoa D. Doan$^1$, Jianwen Xie$^2$, Yaxuan Zhu$^3$, Yang Zhao$^4$, Ping Li$^5$} \\\\
$^1$College of Engineering and Computer Science, VinUniversity 
$^2$Akool Research\\ $^3$Department of Statistics and Data Science, UCLA $^4$Google Research $^5$VecML\\
}

\begin{document}

\maketitle
\begin{abstract}
Leveraging supervised information can lead to superior retrieval performance in the image hashing domain but the performance degrades significantly without enough labeled data. One effective solution to boost performance is to employ generative models, such as Generative Adversarial Networks (GANs), to generate synthetic data in an image hashing model. However, GAN-based methods are difficult to train, which prevents the hashing approaches from jointly training the generative models and the hash functions. This limitation results in sub-optimal retrieval performance. To overcome this limitation, we propose a novel framework, the generative cooperative hashing network, which is based on energy-based cooperative learning. This framework jointly learns a powerful generative representation of the data and a robust hash function via two components: a top-down contrastive pair generator that synthesizes contrastive images and a bottom-up multipurpose descriptor that simultaneously represents the images from multiple perspectives, including probability density, hash code, latent code, and category. The two components are jointly learned via a novel likelihood-based cooperative learning scheme. We conduct experiments on several real-world datasets and show that the proposed method outperforms the competing hashing supervised methods, achieving up to 10\% relative improvement over the current state-of-the-art supervised hashing methods, and exhibits a significantly better performance in out-of-distribution retrieval.

\end{abstract}
\vspace{-5pt}
\section{Introduction} \label{sec:introduction}
\vspace{-5pt}

The rapid growth of digital data, especially images, brings many challenges to the problem of similarity search. A complete linear scan of all the images in such massive databases is computationally expensive, especially when the database contains millions (or billions) of items. To guarantee both computational efficiency and retrieval accuracy, approximate nearest-neighbor (ANN) methods have become increasingly important. The research on efficient algorithms for ANN search dates back  to~\citep{friedman1975algorithm,friedman1977algorithm}. 
Hashing is an ANN method that has several advantages compared to other ANN methods~\citep{broder1997syntactic,indyk1998approximate,charikar2002similarity,datar2004locality,li2005using,lv2007multi,pauleve2010locality,shrivastava2012fast,li2019sign,li2022gcwsnet}. 
In hashing, high-dimensional data points are projected onto a much smaller locality-preserving \textit{binary} space. \textit{Searching for similar images can be efficiently performed} in this binary space using computationally efficient hamming distance. 


\vspace{0.1in}

This paper focuses on the learning-to-hash methods that ``learn'' data-dependent hash functions for efficient image retrieval. Prior studies on learning-to-hash include~\citep{kulis2009learning,grauman2013learning,gong2013iterative,wang2016learning,gui2018r2sdh,Dong2020Learning}. Recently, deep hashing methods learn preserving representations while simultaneously controlling the quantization error of binarizing the continuous representation to binary codes~\citep{lai2015simultaneous,zhu2016deep,cao2017hashnet,yuan2020central,jin2020ssah,hoe2021one,doan2022one}. These methods achieve state-of-the-art performance on several datasets. However, their performance is limited by the amount of supervised information available in the training dataset. Sufficiently annotating massive datasets (common in retrieval applications) is an expensive and tedious task. Subject to such scarcity of supervised similarity information, deep hashing methods run into problems such as overfitting and train/test distribution mismatch, resulting in a significant loss in retrieval performance. 

To improve generalization, ldata-synthesis techniques are proposed for hashing~\citep{qiu2017deep,cao2018hashgan}. These methods rely on increasingly powerful generative models, such as Generative Adversarial Network (GAN), and achieve a substantial retrieval improvement. 
However, training GANs is difficult because of problems such as mode collapse. In existing GAN-based methods, the generative model is first trained,
then the trained generator model is used to synthesize additional data to fine-tune the hash function while the discriminator is usually discarded. This separation is ad hoc and does not utilize the full power of generative models, such as representation learning. 

On the other hand, energy-based models (EBMs), parameterized by modern neural networks for generative learning, have recently received significant attention in computer vision. For example, EBMs have been successfully applied to generations of images \citep{xie2016theory}, videos \citep{xie2017synthesizing}, and 3D volumetric shapes \citep{xie2018learning}.
Maximum likelihood estimation of EBMs typically relies on Markov chain Monte Carlo (MCMC) \citep{liu2008monte,barbu2020monte} sampling, such as Langevin dynamics \citep{neal2011mcmc}, to compute the intractable gradient for updating the model parameters. MCMC is known to be computationally expensive, therefore \citep{XieLGW18,xie2018cooperative,xie2022tale} propose the Generative Cooperative Networks (CoopNets) to learn an EBM, which is called descriptor, together with a GAN-style generator serving as an amortized sampler for efficient learning of the EBM. CoopNets jointly trains the descriptor as a teacher network and the generator as a student network via the MCMC teaching algorithm \citep{XieLGW18}. The generator plays the role of a fast sampler to initialize the MCMC of the descriptor for fast convergence, while the descriptor teaches the generator how to mimic the MCMC transition such that the generator can be a good approximate sampler for the descriptor. 

Compared to GANs, CoopNets' framework has several appealing advantages: First, cooperative learning of two probabilistic models is based on maximum likelihood estimation, which generally does not suffer from GAN's mode collapse issue. Second, while GANs' bottom-up discriminator becomes invalid after training because it fails to tell apart the real and fake examples and only the generator is useful for synthesis purposes, in the CoopNets framework, both bottom-up descriptor and top-down generator are valid models for representation and generation.      
Our paper belongs to the category of energy-based cooperative learning where we bring in the powers of representation and generation of CoopNets to hashing, further advancing the state-of-the-art performance in this domain.

\begin{figure*}[t]
    \centering
    \includegraphics[width=4.0in]{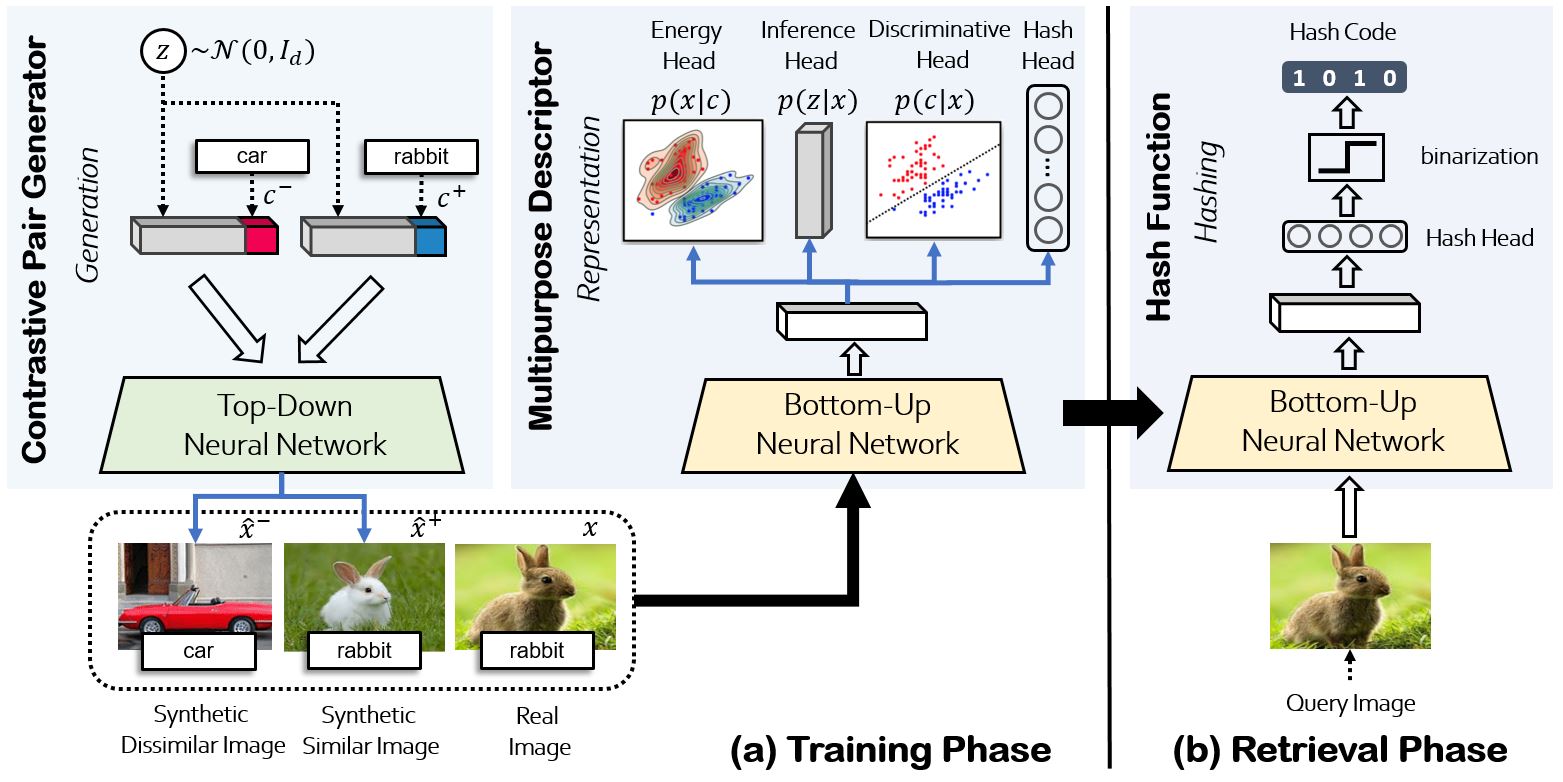}
    \vspace{-10pt}
    \caption{CoopHash consists of two main components: 1) a contrastive pair generator, that takes as inputs the concatenation of a random noise vector $z$ and a label $c^+$ and synthesizes a contrastive image pair $\{\hat{x}^+,\hat{x}^-\}$s from the same class $c^+$ and a different class $c^-$. 2) a multipurpose descriptor that describes the images in multiple ways, including an explicit density model $p(x|c)$, an variational inference model $p(z|x)$, a discriminative model $p(c|x)$, and a hashing model. 
    All four models share a base bottom-up representational network. The multipurpose descriptor network is trained by a loss including negative maximum likelihood, variational loss, triplet-ranking loss, and classification loss, while the contrastive pair generator learns from the descriptor and serves as a fast initializer of the MCMC of the descriptor. In the retrieval phase, only the hashing computational path is used; the binary hash codes are the signs of the real-valued outputs.}
    \label{fig:model}
    \vspace{-15pt}
\end{figure*}

Specifically, building on the foundation of cooperative learning, we jointly train a novel multipurpose descriptor for representation and a contrastive pair generator for generation via MCMC teaching. In the context of cooperative learning with a generator, the proposed multipurpose descriptor simultaneously learns to represent the images from multiple perspectives, including the probability density of the image (with an energy head and a negative log-likelihood loss), the latent code of the image (with an inference head and a VAE loss), the category of the image (with a discriminative head and a classification loss), and especially the hash code of the image (with a hash head and a triplet loss). Our hash function is part of the novel, multipurpose architecture of the descriptor. The modules of this cooperative network play different, but essential roles in learning higher-quality hash codes with desirable properties : (i) The contrastive pair generator learns to synthesize contrastive samples to improve the generalization of the hash function; (ii) The energy head learns effective image representations and improves the robustness of the learned hash function against out-of-distribution (OOD) data; and (iii) The inference head improves the training efficiency of the model and helps recover the corrupted data during retrieval, thus improving the retrieval robustness. 
The proposed CoopHash framework is illustrated in Figure~\ref{fig:model}.

In a controlled environment, the learned hash function achieves state-of-the-art results. When there is a small conceptual drift in the underlying data distribution, a realistic scenario in today's digital world, the learned hash function can still retrieve the most relevant results to the query while in other existing hashing methods, such retrieval performance degrades significantly. Finally, our approach can handle corrupted data in both training and testing, making our method well-suited for real-world applications.
The contributions of our paper are summarized below:
\begin{itemize}[leftmargin=*]
   \item We are the first to study the problem of supervised image hashing in the context of generative cooperative learning, where we specially design a pair of descriptor (energy-based model) and generator (latent variable model), and jointly train them via likelihood-based cooperative training. 

   \item We extend the original cooperative learning framework into a multi-task version by proposing a novel multi-headed or multipurpose descriptor, which integrates cooperative learning (i.e., MCMC teaching process), hash coding (i.e., triplet-ranking loss), classification, and variational inference into a single framework to improve both the generalization capacity of the learned hashing model and the cooperative learning efficiency. 
   \item We train our model (i.e., the descriptor and the generator) in an improved cooperative learning algorithm, where the MCMC-based inference step of the generator in the original cooperative learning framework is replaced by the variational inference for computational efficiency. The resulting training strategy becomes a novel variational MCMC teaching algorithm.  
\item We provide theoretical understanding, including convergence analysis and mode collapse analysis, for our model (please see the appendix). 
   \item We conduct extensive experiments, including the conventional retrieval evaluation and out-of-distribution retrieval evaluation, on several benchmark datasets to demonstrate the advantages of the proposed  framework over several state-of-the-art hashing techniques. 
\end{itemize}
\vspace{-10pt}
\section{Related Work} \label{sec:related}
\vspace{-5pt}
\subsection{Image hashing} 
\vspace{-5pt}

In hashing, shallow methods learn linear hash functions and rely on carefully-constructed features that are extracted from any hand-crafted feature extraction techniques or representation-learning algorithms. Conversely, the deep hashing methods combine the feature representation learning phase and the hashing phase into an end-to-end model and have demonstrated significant performance improvements over the hand-crafted feature-based approaches~\citep{xia2014supervised,cao2017hashnet,doan2022one}. 

\noindent \textbf{Generative Supervised Hashing.} \; Hashing methods can also be divided into unsupervised~\citep{gong2013iterative,weiss2008spectral,yang2019distillhash,dizaji2018unsupervised,lin2016learning,li2022gcwsnet} and supervised hashing~\citep{shen2015supervised,yang2018supervised,ge2014graph,gui2018r2sdh, deng2019two,yuan2020central,jin2020ssah,li2023semantic,lei2023image,xu2021idhashgan,wei2023attribute,zhang2023deep,wang2023deep}. Supervised methods demonstrate superior performance over unsupervised ones, but they can easily overfit when there are limited labeled data. To overcome such limitations, data synthesis techniques have been successfully used in image hashing to improve the retrieval performance~\citep{qiu2017deep,gao2018learning}. These methods employ generative models, such as the popular GAN, to synthesize contrastive images. However, GANs are difficult to train and their generative models do not have any usefulness toward learning the hash functions beyond data synthesizing. 
\vspace{-5pt}
\subsection{Generative cooperative network}
\vspace{-5pt}

\citet{xie2018cooperative,xie2022tale} propose a powerful generative model, called generative cooperative network (CoopNets), which is able to generate realistic image and video patterns. The CoopNets framework jointly trains an EBM (i.e., descriptor network) and a latent variable model (i.e., generator network) via a cooperative learning scheme, where the descriptor is trained by MCMC-based maximum likelihood estimation~\citep{xie2016theory}, while the generator learns from the descriptor and serves as a fast initializer of the MCMC of the descriptor. Other variants include CoopVAEBM \citep{XieZL21} and CoopFlow \citep{XieZLL22}.
~\citet{xie2019cooperative} study the conditional version of CoopNets for supervised image-to-image translation. As to applications, \citet{ZhangXZB22} apply the conditional framework to generative salient object detection. 
\citet{xie2021cycleCoopNets} propose to jointly train two CoopNets models with cycle consistency for unsupervised image-to-image translation. \citet{zhu2023learning} turns the cooperative network into a diffusion generative model. \textbf{Most of the above works focus on leveraging CoopNets for data synthesis.} Our paper studies generative hashing based on the cooperative learning scheme. 
\vspace{-10pt}
\section{Cooperative Hashing Network}
\vspace{-5pt}

The CoopHash framework (described in Figure~\ref{fig:model}) consists of a contrastive pair generator network and a multipurpose descriptor network. They are jointly trained by an MCMC-based cooperative learning algorithm ~\citep{xie2018cooperative}. 
\vspace{-5pt}
\subsection{Problem Statement}
\vspace{-5pt}
Given a dataset $\mathcal{X} = \{x_1,x_2,...,x_n\}$ of $n$ images, 
hashing aims to learn a discrete-output, nonlinear mapping function $\mathcal{H}: x \xrightarrow{} \{-1,1\}^K$, which encodes each image $x$ into a $K$-bit binary vector such that the similarity structure between the images is preserved in the discrete space. In supervised hashing, each example $x_i \in \mathcal{X}$ is associated with a label $c_i$. Note that this is a point-wise label of an image. Another also common supervised scenario has the pairwise similarity label for each pair of images. However, for most image applications, pair-wise labeling is significantly labor-intensive because a dataset of $n$ images requires $n^2$ pairwise labels.  
\vspace{-5pt}
\subsection{Contrastive Pair Generator Network}
\vspace{-5pt}
Let $g(c,z; \Lambda)$ be a nonlinear mapping function parameterized by a top-down decoder network. $\Lambda$ contains all the learning parameters in the network. The conditional generator in the form of a latent variable model is given by
\begin{equation}    
    z \sim \mathcal{N}(0,I_d), x=g(c,z;\Lambda)+ \epsilon, \epsilon \sim \mathcal{N}(0,\sigma^2 I_D), \nonumber
\end{equation}
which defines an implicit conditional distribution of an image $x$ given a label $c$, i.e., $p(x|c;\Lambda)=\int p(z)p(x|c, z;\Lambda)dz$, where $p(x|c,z;\Lambda)=\mathcal{N}(g(x,z;\Lambda), \sigma^2 I_D)$. We further revise the generator for contrastive pair generation given a query example $(c_i, x_i)$ sampled from the empirical data distribution $p_{\text data}(c,x)$. The contrastive pair generator produces a pair of synthetic examples, consisting of one similar example generated with the same label of the query example $c^+=c_i$ and one dissimilar example generated with a different label $c^-\neq c_i$. Both share the same latent code $z$ for semantic feature preservation. To be specific, 
\begin{eqnarray}
\begin{aligned}
   & z \sim \mathcal{N}(0,I_d); \;\; c^+, c^- \sim p_{\text data}(c); \;\; \epsilon \sim \mathcal{N}(0,\sigma I_D); \\
 &   x^+=g(c^+,z;\Lambda)+ \epsilon;\;\;\; x^-=g(c^-,z;\Lambda)+ \epsilon
\end{aligned} \label{eq:contrative_generator}
\end{eqnarray}
The generator plays two key roles in the framework: the first one is to provide contrastive pairs for contrastive learning of the hash function, which is the main goal of the framework, while the second one is to serve as an approximate sampler for efficient MCMC sampling and training of the energy-based descriptor, which is the core step of the cooperative learning. Although the first role aims at the target, i.e., training the hash function, the second role is the foundation of the learning. Without the second role, the generator cannot be trained successfully such that it will fail to generate useful contrastive image pairs.    

In the cooperative training scheme, the generator as an ancestral sampler learns to approximate the MCMC sampling of the EBM. Thus, the learning objective is to minimize the negative log-likelihood of the samples $\{(\tilde{x}_i|c_i)\}_{i=1}^{\tilde{n}}$ drawn from the EBM, i.e., $\mathcal{L}_{G}(\Lambda)=\frac{1}{\tilde{n}} \sum_{i=1}^{n} \log p(\tilde{x}_i|c_i;\Lambda)$,
which amounts to minimizing the following objective
\begin{equation}
\mathcal{L}_{G}(\Lambda)=\frac{1}{n} \sum_{i=1}^{n} ||\tilde{x}_i-g(c_i,\tilde{z}_i; \Lambda)||^2,
\label{eq:generator_loss}
\end{equation}
where $\tilde{z}_i \sim p(z|\tilde{x}_i;\Lambda)$ is the corresponding latent vector inferred from $\tilde{x}_i$. In the original cooperative learning algorithm \citep{xie2018cooperative}, the inference process is typically achieved by MCMC sampling from the intractable posterior  $p(z|\tilde{x}_i;\Lambda)$. In Section \ref{sec:inference}, we propose to learn an encoder as a fast inference model with reparameterization trick \citep{kingma2014auto} to speed up the computation of Eq.~(\ref{eq:generator_loss}). Both the EBM and the inference model are represented by the multipurpose descriptor network. 
\vspace{-5pt}
\subsection{Multipurpose Descriptor Network} 
\vspace{-5pt}

The multipurpose descriptor aims at representing the images from different perspectives. We propose to parameterize the descriptor by a multi-headed bottom-up neural network, where each branch accounts for one different representation of the image. The proposed descriptor assembles four types of representational models of data in a single network in the sense that all models share a base network but have separate lightweight heads built on the top of the base network for different representational purposes. Let $f_0(x;\theta_0)$ be the shared base network with parameters $\theta_0$.   

\noindent\textbf{Energy head.}
The energy head $h_{E}$ along with the base network $f_{0}$ specifies an energy function $f_E(x,c;\Theta_E)$, where observed image-label pairs are assigned lower energy values than unobserved ones. For notation simplicity, let $\Theta_E=(\theta_0,\theta_E)$, and then the energy function $f_{E}(x,c;\Theta_E)=h_{E}(c, f_0(x,\theta_0);\theta_E)$. With the energy function $f_E$, the descriptor explicitly defines a probability distribution of $x$ given its label $c$ in the form of energy-based model
\begin{equation}
p(x|c; \Theta_E) = \frac{p(x,c;\Theta_E)}{\int p(x,c;\Theta_E)dx} = \frac{\exp[-f_{E}(x,c; \Theta_E)]}{Z(c;\Theta_E)}, 
\label{ebm_head}
\end{equation}
where $Z(c;\Theta_E)=\int \exp[-f_{E}(x,c; \Theta_E)]dx$ is the intractable normalizing constant. Eq.~(\ref{ebm_head}) is also called generative modeling of neural network $f_{E}$ \citep{xie2016theory}. The training of $\theta_E$ in this context can be achieved by maximum likelihood estimation, which will lead to the ``analysis by synthesis'' algorithm \citep{grenander2007pattern}. Given a set of training images with labels $\{(c_i,x_i)\}_{i=1}^n$,
we train $\Theta_{E}$ by minimizing the negative log-likelihood (NLL):
\begin{equation}
\mathcal{L}_{\text{NLL}}(\Theta_E) = -\frac{1}{n} \sum_{i=1}^{n} \log p(x_i|c_i;\Theta_E),
\end{equation}
whose gradient is given by
\begin{equation}
\frac{1}{n} \sum_{i=1}^{n} \lbrace \frac{\partial f_{E}(x_i,c_i;\Theta_E)}{\partial \Theta_E} -  \mathbb{E}_{p(x|c_i;\Theta_E)}[\frac{\partial f_{E}(x,c_i;\Theta_E)}{\partial \Theta_E} ]\rbrace, \nonumber
\end{equation}
where the $\mathbb{E}_{p(x|c_i;\Theta_E)}$ denotes the intractable expectation with respect to $p(x |c_i;\Theta_{E})$, which can be approximated by the average of MCMC samples.  We can rely on a cooperative MCMC sampling strategy that draws  samples by (i) first generating initial examples $\hat{x}$ by the generator, and then (ii) revising $\hat{x}$ by finite steps of Langevin updates \citep{zhu1998grade} to obtain final $\tilde{x}$, that is 
\begin{align}
& \hat{x}= g(c,\hat{z};\Lambda), \hat{z} \sim \mathcal{N}(0, I_{d}) \label{eq:MCMC_coop1} \\
& \tilde{x}_{t+1} = \tilde{x}_{t} - \frac{\delta^2}{2} \frac{\partial f_E(\tilde{x}_{t},c;\Theta_E)}{\partial \tilde{x}} + \delta \mathcal{N}(0,I_D), 
\label{eq:MCMC_coop2}
\end{align}
where $\tilde{x}_{0}=\hat{x}$, $t$ indexes the Langevin time steps, and $\delta$ is the step size. The Langevin dynamics in Eq.~(\ref{eq:MCMC_coop2}) is a gradient-based MCMC, which is equivalent to a stochastic gradient descent algorithm that seeks to find the minimum of the objective function defined by $f_E(x,c;\Theta_E)$. Eq.~(\ref{eq:MCMC_coop1}) is accomplished by the generation process presented in Eq.~(\ref{eq:contrative_generator}). With the MCMC examples $\{(\tilde{x}_i|c_i)\}$, we compute the gradient by
\begin{equation}
     \nabla(\Theta_E) \approx \frac{1}{n} \sum_{i=1}^{n} \left[\frac{\partial f_E(x_i,c_i;\Theta_E)}{\partial \Theta_E} - \frac{\partial f_E(\tilde{x}_i,c_i;\Theta_E)}{\partial \Theta_E}\right] \nonumber
\label{equ:ebm_update}
\end{equation}
where $\frac{\partial f_E}{\partial \Theta_E}$ can be efficiently computed by backpropagation.

\noindent\textbf{Inference head.} \label{sec:inference}
With the reparameterization trick, the descriptor can play the role of the inference model to assist the training of the generator. The inference model, which approximates the intractable posterior distribution $p(z|x,c;\Lambda)$ of the generator,  can be parameterized by a multivariate Gaussian with a diagonal covariance structure
 $\pi(z|x,c; \Theta_I) \sim \mathcal{N}(\mu(x,c;\Theta_I), {\rm diag}(v(x,c;\Theta_I)))$, where $\mu(x,c;\Theta_I)=\mu(f_{0}(x;\theta_0),c;\theta_I)$ and $v(x,c;\Theta_I)=v(f_{0}(x;\theta_0),c;\theta_I)$ are $d$-dimensional outputs of encoding networks of image $x$ conditioned on $c$, with trainable parameters $\Theta_{I}=(\theta_0,\theta_{I})$. 
 
 The cooperative learning of the generator needs to infer the latent variables $z$ by using an iterative MCMC, such as Langevin dynamic, to compute the intractable posterior distribution $p(z|x,c;\Lambda)=p(x|z,c;\Lambda)p(z)/p(x|c;\Lambda)$. The inference step is important but computationally expensive because it might take a long time for MCMC to converge. In our framework, the amortized inference model $\pi(z|x,c;\Theta_{I})$ learns an efficient mapping from image space $x$ to latent space $z$ to approximate the posterior distribution in the context of variational auto-encoder (VAE) \citep{kingma2014auto}, and takes the place of the MCMC-based inference procedure in the original cooperative training algorithm. 
 
Specifically, the generator $p(x|c,z;\Lambda)$ and the inference model $\pi(z|x,c; \Theta_I)$ forms a special VAE that treats the MCMC examples $\{(\tilde{x}_i,c_i)\}_{i=1}^{\tilde{n}}$ generated from the EBM $p(x|c;\theta_{E})$ as training examples. This will lead to a variational MCMC teaching algorithm \citep{xie2021learning} that combines the cooperative learning and variational inference.
We can learn $\alpha$ and $\Theta_I$ by minimizing the variational lower bound of the negative log-likelihood of  $\{(\tilde{x}_i,c_i)\}_{i=1}^{\tilde{n}}$:
\begin{eqnarray} 
\begin{aligned} 
\mathcal{L}_{\text{VAE}}(\Lambda, \Theta_{I}) &= \frac{1}{\tilde{n}} \sum_{i=1}^{\tilde{n}} [ -\log p(\tilde{x}_i|c_i;\Lambda) + \gamma \text{KL}(\pi(z_i|\tilde{x}_i,c_i;\Theta_{I})||p(z_i|\tilde{x}_i,c_i;\Lambda))],
\label{eq:vae_lower_bound}
\end{aligned} 
\end{eqnarray} 
where $\gamma$ is a hyperparameter that controls the relative importance between two terms. Different from \citep{xie2021learning}, our framework is a conditional model that takes both the image and label as inputs, and the EBM and inference model in our framework have a shared base network. 

\noindent\textbf{Hash head.} The descriptor, with the hash head $h_{H}$, learns to represent the input images as binary hash codes. The hash head $h_{H}$ and the base network $f_{0}$ form a hash function $f_{H}(x;\Theta_H)=h_{H}(f_{0}(x;\theta_0);\theta_H)$, where $\Theta_H=(\theta_0,\theta_H)$. The hash function aims to preserve the semantic similarity of the images in the discrete space. One effective way to learn such a hash function is: for each image $x$, we ``draw'' a positive sample $x^+$ which is conceptually similar to $x$ and a negative sample $x^-$ which is conceptually dissimilar to $x$, and train the hash function to produce similar hash codes for $x$ and $x^+$, and dissimilar hash codes for $x$ and $x^-$. Such contrastive learning can be achieved by recruiting the contrastive pair generator in Eq.~(\ref{eq:contrative_generator}), with which we can synthesize the negative and positive examples by sampling from the class-conditional distribution. For each observed image $x$ and its label $c$, we can easily generate a synthetic image $x^+$, conditioned on the label $c$, and a synthetic image $x^-$, conditioned on a different label $c^{-1} \ne c$. The three examples form a real-synthetic triplet $(x, x^+, x^-)$. The hash function $f_{H}$ can be trained to minimize the Hamming distance (a discrete distance function which is typically approximated by the continuous $L_2$ distance) between $f_{H}(x)$ and $f_{H}(x^+)$ and maximize the distance between $f_{H}(x)$ and $f_{H}(x^-)$. This triplet-ranking loss is defined as follows:
\begin{eqnarray}
\begin{aligned}
\label{eqn:l_h_relaxed}
 \mathcal{L}_{\text{TR}} (\Theta_H) =&||f_{H}(x)-f_{H}(x^+)||_2 + \max(m- ||f_{H}(x)-f_{H}(x^-)||_2, 0)  \\ 
& + \lambda (|||f_{H}(x)|-1||_2 + |||f_{H}(x^+)|-1||_2 + |||f_{H}(x^-)|-1||_2) \nonumber
\end{aligned}
\end{eqnarray}
where $|.|$ is element-wise absolute operation. The first term preserves the similarity between images with similar concepts, while the second term punishes the mapping of semantically dissimilar images to similar hash codes if their distance is within a margin $m$. The last term minimizes the quantization error of approximating the discrete solution with the real-value relaxation.

\noindent\textbf{Discriminative head.}\   Image labels provide not only knowledge for classification but also supervised signals for high-level image understanding. On the other hand, the hash codes should be predictive of the image labels. This relationship can be modeled through a multi-class classification problem. For each $x_i$, let $\hat{c}_i$ be the predicted label.
The multi-class classification loss can be defined as:
\begin{equation}
\mathcal{L}_{\text{CLASS}}(\Theta_C) = - \frac{1}{n}\sum_{i=1}^{n} c_i \log \frac{e^{\theta_{c_i,:}^T h(x)}}{\sum_{j} e^{\theta_{j,:}^T h(x_i)}} \label{eqn:l_c}
\end{equation}
where  $\theta_C \in \mathbb{R}^{K \times L}$ is the parameter of the linear layer that maps each hash code into the class labels. 
\vspace{-5pt}
\subsection{Optimization}
\vspace{-5pt}
The proposed cooperative hashing network jointly trains the contrastive pair generator network and the multipurpose descriptor network. At each iteration, the EBM $p(x|c;\Theta_E)$ in the descriptor and the contrastive pair generator $p(x|c;\theta)$ produce synthetic contrastive image pairs by following the cooperative sampling strategy described in Eq.~(\ref{eq:MCMC_coop1}) (or more specifically, Eq.~(\ref{eq:contrative_generator})) and Eq.~(\ref{eq:MCMC_coop2}). With the synthetic images, we can update the generator and the descriptor respectively by minimizing the following losses:

\noindent\textbf{Descriptor loss.}\ We train the multipurpose descriptor to simultaneously describe the images from multiple representational perspectives. The overall training objective of the descriptor, which combines the negative log-likelihood $\mathcal{L}_{\text{NLL}}(\Theta_E)$, the variational loss $\mathcal{L}_{\text{VAE}}(\Theta_I)$ ,the triplet-ranking loss $\mathcal{L}_{\text{TR}}(\Theta_H)$, and the classification loss $\mathcal{L}_{\text{CLASS}}(\Theta_C)$, is given by
\begin{eqnarray}
\begin{aligned}
\mathcal{L}_{\text{Des}} (\Theta_*)
 = \mathcal{L}_{\text{NLL}}(\Theta_E) + \beta_{I} \mathcal{L}_{\text{VAE}}(\Theta_I)  + \beta_{H} \mathcal{L}_{\text{TR}}(\Theta_H) + \beta_{C} \mathcal{L}_{\text{CLASS}}(\Theta_C).
\end{aligned}
\label{eq:des_loss}
\end{eqnarray}
where $\beta_*$ are the parameters to balance the weights between different losses.

\noindent\textbf{Generator loss.}\  For the generator, we optimize:
\begin{equation}
\mathcal{L}_{\text{Gen}} (\Lambda) = \mathcal{L}_{\text{VAE}}(\Lambda)
\label{eq:gen_loss}
\end{equation}
After training, the real-valued hash code of an image is the output of $f_H$, and its discrete version is given by $\mathcal{H} = sign \circ f_H$. The Algorithm is provided in Supplement.

\noindent\textbf{Multi-task learning.}\   Our work belongs to multi-task learning \citep{caruana1997multitask}, in which multiple learning tasks are solved in parallel. Specifically, our framework simultaneously learns to perform three tasks: (i) \textit{Hashing} (the main task): contrastive learning of a hash function with the help of a contrastive pair generator; (ii) \textit{Image generation}: learning a class-conditional image generative model using cooperative learning, where an EBM, a generator model, and an inference model are jointly trained by variational MCMC teaching; (iii) \textit{Image classification}: supervised learning of a classifier for a discriminative task. These three tasks are not independent. The generator plays the role of the approximate sampler in task (ii), while serving as a provider of synthetic contrastive pairs in task (i). Besides, the EBM, the inference model, the discriminative model, and the hash function share parameters in their base network, therefore their learning behaviors influence one another. (For convergence analysis, please refer to Section \ref{sec:convergence_analysis} in Appendix.)    
\vspace{-10pt}
\section{Experiments}
\vspace{-5pt}
We present evaluation results on several benchmark datasets to demonstrate CoopHash's effectiveness.
\vspace{-10pt}
\subsection{Experimental setup}
\vspace{-5pt}
\noindent \textbf{Dataset.}\ We evaluate the methods on 3 popular datasets in the image hashing domain: \textbf{NUS-WIDE} (269,648 images and 81 labels), \textbf{COCO} (123,287 images and 80 labels) and \textbf{CIFAR-10} (60,000 images and 10 labels). 
In \textbf{NUS-WIDE} and \textbf{COCO}, each image can be tagged with multiple labels, while in \textbf{CIFAR-10}, each image has a single label.

\noindent \textbf{Metrics:} We evaluate the retrieval performance of the methods using the standard retrieval metrics: Mean Average Precision at $k$ (mAP@$k$) and Precision at $k$ (P@$k$). Besides reporting P@1000, we calculate mAP@54000 for CIFAR-10 and mAP@5000 for NUS-WIDE and COCO. 


\noindent\textbf{Baselines.}\ We compare CoopHash against representative approaches, including the shallow methods (BRE~\citep{kulis2009learning}, ITQ~\citep{gong2013iterative}, KSH~\citep{liu2012supervised}, and SDH~\citep{shen2015supervised}), the deep methods (CNNH~\citep{xia2014supervised}, DNNH~\citep{lai2015simultaneous}, DHN~\citep{zhu2016deep}, DSDH~\citep{li2017deep}, HashNet~\citep{cao2017hashnet}, DCH~\citep{cao2018deep}, GreedyHash~\citep{su2018greedy}, DBDH~\citep{zheng2020deep}, CSQ~\citep{yuan2020central}, OrthoHash~\citep{hoe2021one}, CSQ-$L_{\text{CPQ}}$~\cite{wang2023deep}, and DCGH~\cite{zhang2023deep}), and the previously proposed synthesis methods, DSHGAN~\citep{qiu2017deep}, HashGAN~\citep{cao2018hashgan}, Idhashgan~\cite{xu2021idhashgan}, and SemGAN~\cite{li2023semantic}). Note that, some deep methods such as GreedyHash or OrthoHash are orthogonal to generative hashing methods such as DSHGAN, HashGAN and our CoopHash
, and can be combined with ours to further improve the retrieval performance.
Additionally, we include the HashGAN variant (HashGAN-1) that jointly learns the generative model and hash function in one stage. 


\textbf{Implementation.} For fair evaluation, we use the same hash-function backbone and input resolution of $32\times32$ for the baselines and CoopHash. Consequently, the reported performance can be slightly different from those in the respective papers. Details are provided in Supplementary.

\begin{table*}[!t]
\setlength{\tabcolsep}{2pt}
\footnotesize
    \centering
    \caption{Mean Average Precision (mAP) for different numbers of bits.}
    \vspace{-10pt}
    \begin{tabular}{l|cccc|cccc|cccc}
        \specialrule{.1em}{.1em}{.1em} 
        
        \multirow{2}{*}{Method} & \multicolumn{4}{|c}{NUS-WIDE} & \multicolumn{4}{|c}{CIFAR-10} & \multicolumn{4}{|c}{COCO} \\
        \cline{2-13}
        & 16 bits & 32 bits & 64 bits & 128 bits & 16 bits & 32 bits & 64 bits & 128 bits & 16 bits & 32 bits & 64 bits & 128 bits  \\ 
        
        \hline
            
        ITQ & 0.460 & 0.405 & 0.373 & 0.347 & 0.354 & 0.414 & 0.449 & 0.462 & 0.566 & 0.562 & 0.530 & 0.502 \\ 
        
        BRE & 0.503 & 0.529 & 0.548 & 0.555 & 0.370 & 0.438 & 0.468 & 0.491 & 0.592 & 0.622 & 0.630 & 0.634 \\ 
        
        KSH & 0.551 & 0.582 & 0.612 & 0.635 & 0.524 & 0.558 & 0.567 & 0.569 & 0.521 & 0.534 & 0.534 & 0.536 \\ 
        
        SDH & 0.588 & 0.611 & 0.638 & 0.667 & 0.461 & 0.520 & 0.553 & 0.568 & 0.555 & 0.564 & 0.572 & 0.580 \\ 
        
        CNNH & 0.570 & 0.583 & 0.593 & 0.600 & 0.476 & 0.472 & 0.489 & 0.501 & 0.564 & 0.574 & 0.571 & 0.567 \\ 
        
        DNNH & 0.598 & 0.616 & 0.635 & 0.639 & 0.559 & 0.558 & 0.581 & 0.583 & 0.593 & 0.603 & 0.605 & 0.610 \\ 
        
        DHN & 0.637 & 0.664 & 0.669 & 0.671 & 0.568 & 0.603 & 0.621 & 0.635 & 0.677 & 0.701 & 0.695 & 0.694 \\ 
        
        DSDH & 0.650 & 0.701 & 0.705 & 0.709 & 0.655 & 0.660 & 0.682 & 0.687 & 0.659 & 0.688 & 0.710 & 0.731 \\ 
        
        HashNet & 0.662 & 0.699 & 0.711 & 0.716 & 0.643 & 0.667 & 0.675 & 0.687 & 0.687 & 0.718 & 0.730 & 0.736 \\ 
        
        DCH & 0.679 & 0.692 & 0.691 & 0.684 & 0.629 & 0.647 & 0.643 & 0.636 & 0.506 & 0.516 & 0.516 & 0.523 \\
        
        GreedyHash & 0.665 & 0.692 & 0.708 & 0.725 & 0.613 & 0.629 & 0.646 & 0.653 & 0.464 & 0.509 & 0.544 & 0.574 \\

        DBDH & 0.690 & 0.702 & 0.716 & 0.725 & 0.575 & 0.577 & 0.588 & 0.594 & 0.511 & 0.519 & 0.526 & 0.538 \\
        
        CSQ & 0.701 & 0.713 & 0.720 & 0.729 &  0.646 & 0.699 & 0.709 & 0.712 & 0.679 & 0.699 & 0.714 & 0.725 \\        

        CSQ-$L_{\text{CPQ}}$ & 0.749 & 0.753 & 0.784 & 0.783 & 0.745 & 0.753 & 0.779 & 0.774 & 0.749 &  0.757 &  0.768 &  0.775 \\

        DCGH & 0.750 & 0.765 & 0.780 & 0.796 & 0.719 & 0.766 & 0.784 & 0.802 & 0.751 &  0.762 &  0.771 &  0.783 \\
        
        \hline
        DSHGAN & 0.721 & 0.745 & 0.767 & 0.782 & 0.701 & 0.753 & 0.764 & 0.760 & 0.730 & 0.742 & 0.757 & 0.764 \\
        
        HashGAN-1 & 0.510 & 0.522 & 0.540 & 0.545 & 0.486 & 0.495 & 0.501 & 0.513 & 0.489 & 0.491 & 0.512 & 0.539 \\
        
        HashGAN & 0.715 & 0.737 & 0.744 & 0.748 & 0.668 & 0.731 & 0.735 & 0.749 & 0.697 & 0.725 & 0.741 & 0.744  \\
        Idhashgan & 0.732 & 0.741 & 0.758 & 0.790 & 0.705 & 0.709 & 0.713 & 0.721 & 0.735 & 0.751 & 0.762 & 0.781 \\
        SCGAH & 0.739 & 0.749 & 0.764  & 0.792 & 0.710 & 0.759 & 0.780 & 0.795 & 0.755 & 0.760 & 0.769 & 0.785 \\
        
        CoopHash & \textbf{0.758} &\textbf{0.771} & \textbf{0.793} & \textbf{0.805} & \textbf{0.727} & \textbf{0.772} & \textbf{0.795} & \textbf{0.809} & \textbf{0.760} & \textbf{0.775} & \textbf{0.780} & \textbf{0.798}  \\ 
        \specialrule{.1em}{.1em}{.1em} 
    \end{tabular}
    \label{tab:result_map}
    \vspace{-10pt}
\end{table*}

\begin{figure}[t]
    \centering
    \begin{minipage}{0.45\textwidth}
        \setlength{\tabcolsep}{1pt}
    \footnotesize
    \caption{mAP performance of OOD Re-\\trieval experiments.}
    \resizebox{0.9\textwidth}{!}{
    \begin{tabular}{lcccc}
        \specialrule{.1em}{.1em}{.1em} 
        Train/\textit{Test}  &  HashNet & CSQ  & HashGAN  & CoopHash \\ \hline
        SVHN/\textit{MNIST} & 0.181 & 0.517 & 0.354 &  \textbf{0.594} \\
        SVHN/\textit{SVHN} & 0.837 & 0.854 & 0.889 & \textbf{0.902} \\
        \hline
        MNIST/\textit{SVHN} & 0.193 & 0.273 & 0.280 & \textbf{0.334} \\
        MNIST/\textit{MNIST} & 0.957 & 0.991 & 0.990 & \textbf{0.995} \\
        \specialrule{.1em}{.1em}{.1em} 
    \end{tabular}
    }
    \label{tab:result_ood}
    \end{minipage}\hfill
    \begin{minipage}{0.55\textwidth}
        \centering
        \includegraphics[width=1.4in]{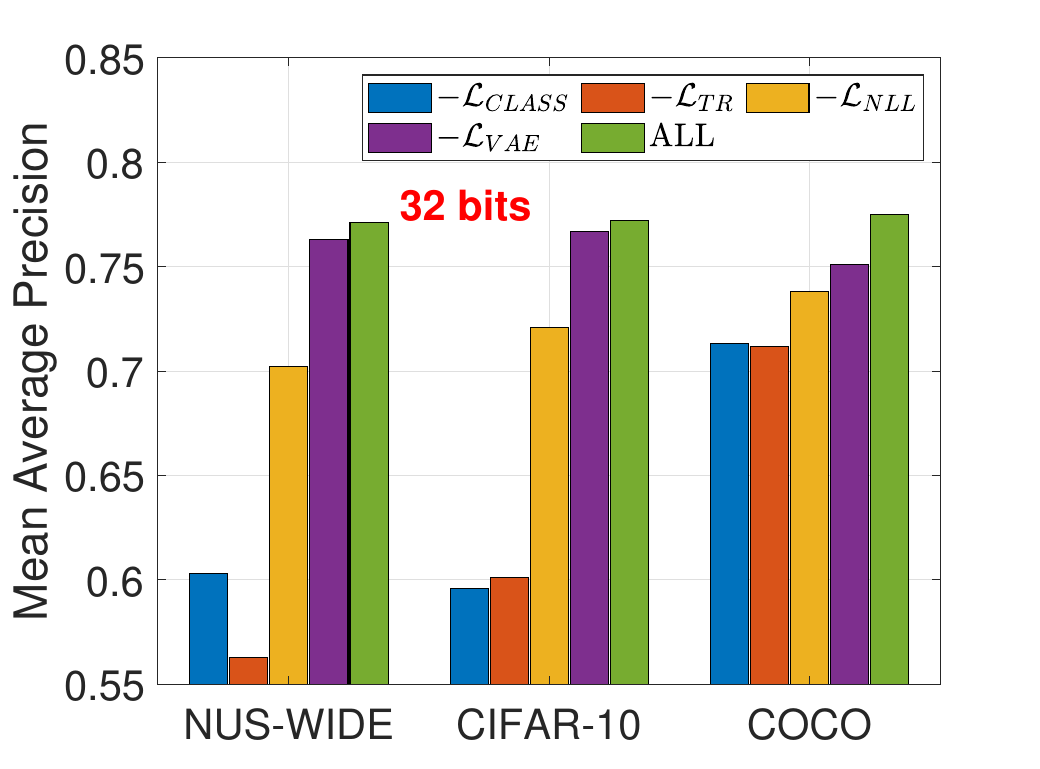}
        \includegraphics[width=1.4in]{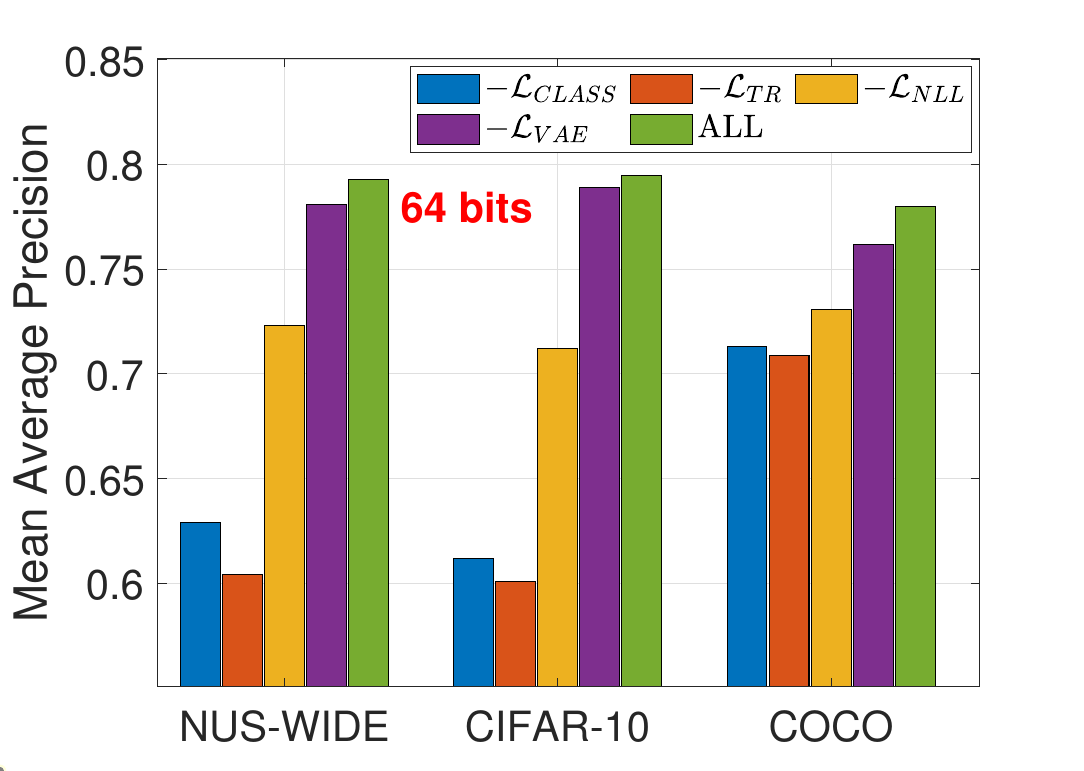}
        \caption{Ablation Study. $-\mathcal{L}_{\text{CLASS}}$: CoopHash without Discriminative head. $-\mathcal{L}_{\text{TR}}$: CoopHash without Hash head. $-\mathcal{L}_{\text{NLL}}$: CoopHash without Energy head.  $-\mathcal{L}_{\text{VAE}}$: CoopHash without Inference head. ALL: CoopHash.}
        \label{fig:result_ablation}
    \end{minipage}
    \vspace{-20pt}
\end{figure}
\vspace{-5pt}
\subsection{Retrieval Results}~\label{sec:main_retrieval_results}
In this section, we present the results of querying for similar images. Table~\ref{tab:result_map} shows the mAP results for all the methods. CoopHash significantly outperforms the shallow methods (by at least 15\% on all datasets). CoopHash also outperforms all deep methods, including the recently proposed DBDH and CSQ, by at least 5-8\%. Finally, CoopHash's performance is better than HashGAN by at least 4\%. The results are statistically significant (with $p$-value $<$ 0.01) on all the datasets. Similar superior performance of CoopHash can be observed on the P@1000 metric in the Supplementary (Table \ref{tab:result_precision}). This makes CoopHash also desirable for practical, precision-oriented retrieval systems. 


The results provide empirical evidence to support the discussion in Section~\ref{sec:introduction}. \textbf{First}, generating synthetic data improves the performance of a supervised method when the labeled data is limited. The generative model improves the amount of ``labeled'' training data and increases its diversity, both of which improve the method's generalization capacity. \textbf{Second}, HashGAN-1 performs significantly worse than HashGAN and CoopHash, indicating the difficulties of training an end-to-end, one-stage data-synthesis hashing based on GANs. \textbf{Finally}, CoopHash, an end-to-end generative multipurpose model, has better performance than the GAN-based methods, demonstrating the advantage of CoopHash's maximum likelihood training and the descriptor's effectiveness in representation learning.

\vspace{-10pt}
\subsection{Retrieval with Out-of-distribution Data}
\vspace{-5pt}
We empirically show that the generative energy-based hashing method exhibits better OOD retrieval than the other models, including the GAN-based approach. In today's digital world where data is continuously added with a high likelihood for conceptual drifts, the method's robustness to such distributional changes is very important. We propose to construct the OOD experiments as follows. In the learning phase, each hashing method is trained using a \textit{train} dataset. In the evaluation phase, we use a different but similar \textit{test} dataset. The test dataset is similar to the train dataset but comes from a (slightly) shifted data distribution. Specifically, we construct two experiments: the first experiment has the train dataset as MNIST and the test dataset as SVHN, and the second experiment has the reverse. 
Table~\ref{tab:result_ood} shows the retrieval results of CoopHash, the state-of-the-art deep methods HashNet and CSQ, and   HashGAN.

        
CoopHash significantly outperforms all the methods in OOD retrieval, with more than 24\% when using MNIST for querying, and 5\% when using SVHN for querying. When being trained on a more complex dataset (SVHN), the retrieval performance of CoopHash significantly improves in the OOD tests, while the OOD performance of the other models only slightly improves. We believe that the EBM descriptor has a flexible structure that allows the learning of the latent representation much more effectively. 

\vspace{-5pt}
\subsection{Ablation Study}
\vspace{-5pt}

We study the contribution of each proposed objective in CoopHash (i.e. the heads). Figure~\ref{fig:result_ablation} shows the mAP values of CoopHash with different combinations of the objectives. As can be observed, each head plays an important role  in improving the retrieval performance. Only training the hash function without data synthesis ($-\mathcal{L}_{NLL}$) results in a significant performance decrease.
Without the triplet ranking loss, i.e. Hash head ($-\mathcal{L}_{\text{TR}}$), the performance also drops significantly. Note that this CoopHash variant is not a hashing method.
The addition of the Discriminative head ($-\mathcal{L}_{\text{NLL}}$) further improves the retrieval performance because the model is trained to preserve the high-level concepts in the hash codes. Finally, the role of the Inference head ($\mathcal{L}_{\text{ALL}}$) is important when it helps achieve both the best performance in CoopHash and faster convergence to the optimal performance. \textbf{The experiment shows that each objective of CoopHash contributes toward CoopHash's optimal retrieval performance and efficient training.}

\vspace{-5pt}
\subsection{Analysis of the Optimal Descriptor}
\vspace{-5pt}






 \begin{wrapfigure}{r}{0.58\textwidth}
\vspace{-15pt}
\centering 
\includegraphics[width=3.4in]{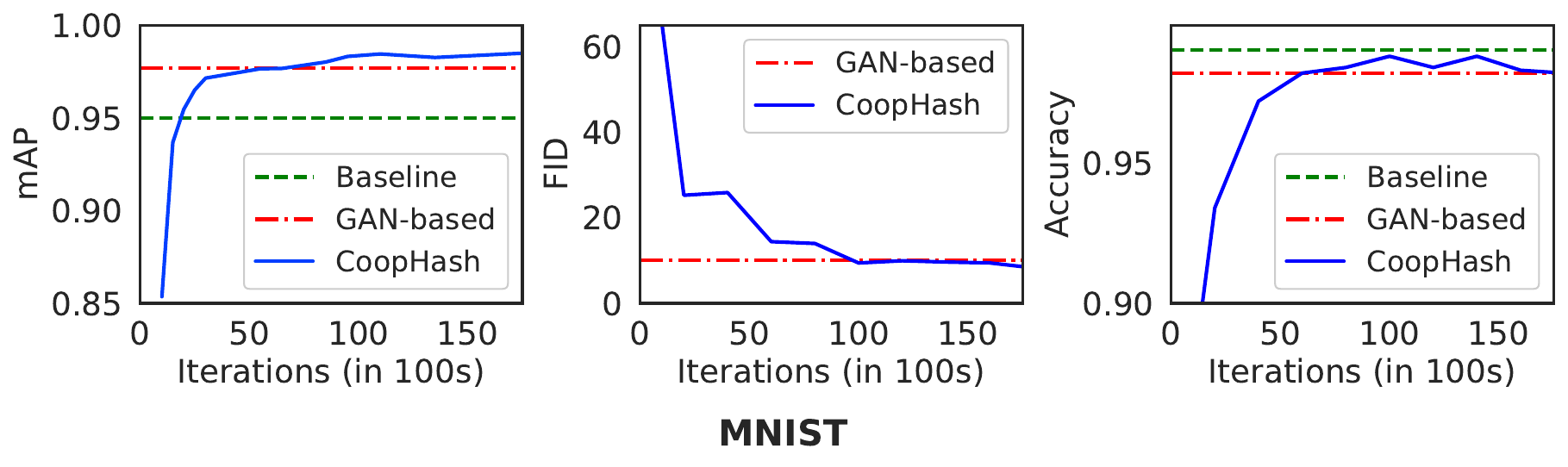}
\includegraphics[width=3.4in]{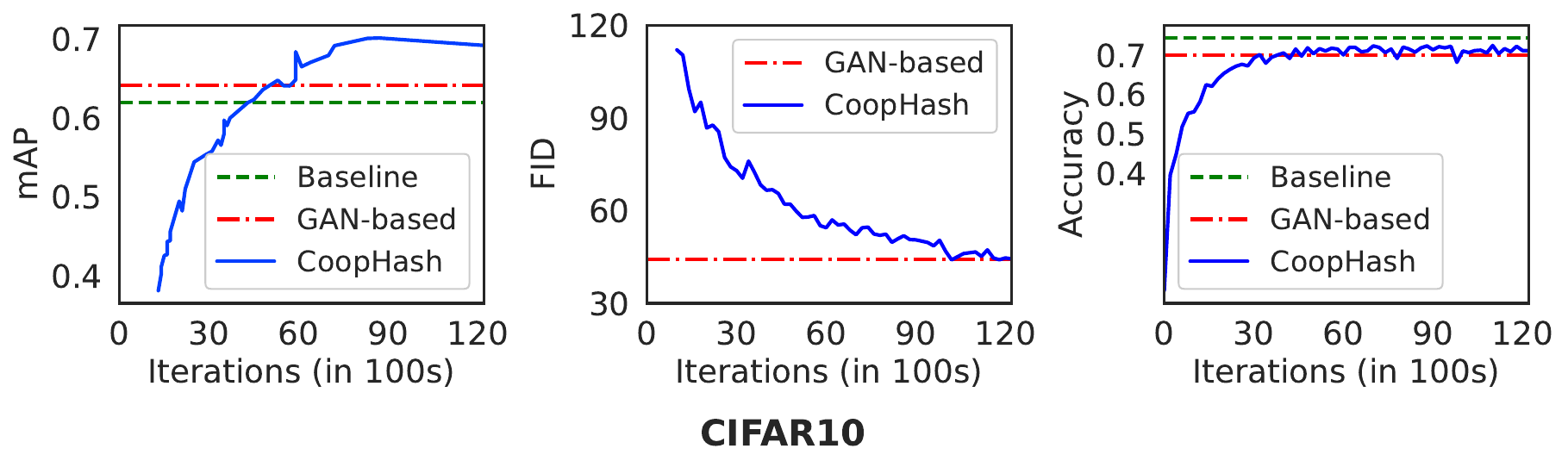}
\vspace{-10pt}
\caption{mAP, FID, and Classification Accuracy of different heads during model training on MNIST and CIFAR10 datasets. In each task, \textbf{Baseline} is a model with similar architecture that is only trained to perform a specific task (i.e. hashing or classification). \textbf{GAN-based}: a similar GAN-based model to CoopHash.}\label{fig:result_optimality}
\vspace{-15pt}
\end{wrapfigure}
We show that the learned representation is simultaneously optimal for the individual tasks solved by EBM, Hash, and Discriminative heads. Figure~\ref{fig:result_optimality} shows the evaluation metrics, including mAP (to evaluate the Hash head), FID (to evaluate the Generator/EBM Head), and Accuracy (to evaluate the Discriminative Head), on the test sets during training on MNIST and CIFAR10. Note that the training sets only have a limited number of 5,000 samples. We can observe that as the training progresses, the quality of the generated images becomes better (i.e. FID decreases). As this happens, the Discriminative Head's Accuracy also increases and approaches the optimal value of the model (with a similar architecture) that is only trained to classify images (\textbf{Baseline}). 

Importantly, the retrieval performance of the Hash Head also increases during training and finally becomes significantly better than the mAP performance of the model that is only trained with the triplets sampled from the real training data (\textbf{Baseline}).
Again, we can see that CoopHash has better generalization than classic deep hashing models with limited labeled data. We also report the optimal performance of the similar \textbf{GAN-based} model where the cooperative objectives $\mathcal{L}_G$ and $\mathcal{L}_{NLL}$ of the generator and descriptor are replaced with the GAN losses~\citep{goodfellow2020generative}. While CoopHash and the GAN-based model have similar image generation quality, CoopHash clearly performs significantly better than the GAN-based model, especially in retrieval.

Note that CoopHash Accuracy is slightly worse than that of the specialized Baseline model because we only synthesize data to improve the Hash Head.   
\vspace{-10pt}
\section{Conclusions}
\vspace{-5pt}
This paper proposes CoopHash, a novel energy-based generative hashing framework, taking full advantage of generative models to learn a high-performing and robust hash function. CoopHash simultaneously learns a contrastive pair generator to synthesize contrastive images and learns a multipurpose descriptor to represent images from multiple perspectives, including classification, hashing, inference, and probability density. The generator and descriptor are trained in a novel MCMC-based cooperative learning scheme. CoopHash learns high-quality binary hash codes and is significantly more robust to out-of-distribution and data-corruption retrieval than the~existing~methods.

\bibliographystyle{abbrvnat}
\bibliography{neurips_2024}


\appendix
\section*{Appendix}
We provides additional details, analysis and experimental results to support the main submission. We begin by providing convergence analysis in Section~\ref{sec:convergence_analysis}, analysis on mode collapse in Section~\ref{sec:mode_collapse}, additional analysis of the classification objective in Section~\ref{sec:coophash_mi}. Then we provide the detail of CoopHash's training algorithm in Section~\ref{sec:training_algo} and discuss in more details the contributions of each CoopHash's components toward learning a better hash function. Next, we discuss additional related methods to CoopHash in Section~\ref{sec:related_hashing_methods}. Then, we discuss the detailed experimental setup and implementation of the methods in Section~\ref{sec:experiment_details} and provide additional experiments on more related works, the hyperparameters of CoopHash and the image generation objective, CoopHash's robustness against data corruption,  as well as discussing the performance of CoopHash for ``unseen setting'' in Section~\ref{sec:additional_experiments}. Finally, we discuss the limitations of our work in Section~\ref{sec:limitations}.

\section{Convergence Analysis}~\label{sec:convergence_analysis}

\subsection{The conditional EBM $\Theta_E$ and the generator $\Lambda$ form the core of the generative model}

Let $K_{\Theta_E}(x|{x}',c)$ denote the transition kernel from $x'$ to $x$ conditioned on $c$, which corresponds to the stochastic process of the $T$-step Langevin sampling that revise the initial example ${x}'$ to the refined examples $x$. The Langevin sampling is driven by the energy-based model parameterized by $\Theta_E$. 

We can further define $(K_{\Theta_E} p_{\Lambda})(x|c)= \int K_{\Theta_E}(x|x',c)p(x'|c;{\Lambda})dx'$ as the conditional distribution of $x$ given $c$, where $x$ is obtained by running $T$ steps of Langevin update starting from the output of the conditional generator $p(x|c; \Lambda)$. 

Consider $p_{data}(x|c)$ as the true conditional distribution for generating $x$ given class label $c$. The maximum likelihood training of the conditional EBM $p(x|c;\Theta_E)$ is equivalent to a minimization of the KL divergence $KL(p_{data}(x|c)||p(x|c;\Theta_E))$. 

Let us use $j$ to index the learning iteration for model parameters. Given the current conditional generator $p(x|c;{\Lambda})$, the EBM updates its parameters $\Theta_E$ by minimizing
\begin{equation}
    {\Theta_E}^{(j+1)}=\arg \min_{\Theta_E} KL(p_{data}(x|c)||p(x|c;\Theta_E))-KL((K_{{\Theta_E}^{(j)}}p_{\Lambda})(x|c)||p(x|c;\Theta_E)) \label{app:coop_objective}
\end{equation}
which is a generalization of the contrastive divergence loss. (It is worth noting that, in the original contrastive divergence, the above $(K_{{\Theta_E}^{(j)}}p_{\Lambda})(x|c)$ is replaced by $(K_{\Theta_E^{(j)}}p_{data})(x|c)$. That is, the MCMC chains are initialized by the true data distribution $p_{data}(x|c)$ instead of the generator $p(x|c;\Lambda)$). The learning shifts  $p(x|c;\Theta_E)$ toward the true distribution $p_{data}(x|c)$. 

The learning of the generator via the original MCMC teaching maximizes the likelihood of the conditional EBM's synthesized examples:
\begin{equation}
L(\Lambda)= \frac{1}{n}\sum_{i=1}^{n} \log p(\tilde{x}_i|c_i;\Lambda),
\end{equation}

where $\tilde{x}$ is the generated examples produced by the conditional EBM. That is, $\tilde{x}_i \sim (K_{\Theta_E}p_{\Lambda})(x|c_i)$.

Thus, given the current EBM $\Theta_E$, the learning of the generator $\Lambda$ approximately follows  
\begin{equation}
{\Lambda}^{(j+1)}=\arg \min_{\Lambda} KL((K_{{\Theta_E}}q_{\Lambda^{(j)}})(x|c)||q_{\Lambda}(x|c)). 
\end{equation}
The learning shift the conditional generator $p(x|c; \Lambda)$ toward the conditional EBM $p(x|c;\Theta_E)$.

As a result, the conditional generator chases the conditional EBM towards the true data distribution. Such a chasing game ensures a maximum likelihood learning of both the conditional generator $q_{\Lambda}(x|c)$ and the EBM $p_{\Theta_E}(x|c)$ (bottom-up network with the energy head).

\subsection{The inference model plays the role of amortized inference for the posterior sampling}

Since the generator is a top-down latent variables model, $p(x|c; \Lambda)$ is an implicit probability density given by $p(x|c; \Lambda) = \int p(x|z,c; \Lambda) p(z) dz$,
where $p(z)$ is Gaussian distribution and $p(x|z,c; \Lambda)$ is also a Gaussian distribution with its mean as $x=g(c,z;\Lambda)$.

Maximizing $L(\Lambda)$ requires an inference step of the latent variables $z$ for each $\tilde{x}$. This typically needs a Langevin sampling that draws samples from the posterior distribution $p(z|x,c; \Lambda)$, and then we can update $\Lambda$ by minimizing the reconstruction loss
\begin{equation}
L(\Lambda) = \frac{1}{n} \sum_{i=1}^{n} || \tilde{x}_i-g(c_i,\tilde{z}_i;\Lambda) ||^2, \quad
\text{where} \quad \tilde{z}_i \sim p(z|\tilde{x}_i,c_i;\Lambda).
\end{equation}
Introducing a bottom-up inference network $\pi(z|x,c;\Theta_I)$ can make the inference step more efficient. Training an approximate inference model for amortizing the posterior sampling requires variational inference technique, in which we learn $\Theta_I$ to minimize the KL divergence between the approximate inference and the posterior distribution $KL(\pi(z|x,c;\Theta_I)||p(z|x,c;\Lambda))$.    

Overall, we can group the conditional inference model $\pi(z|x,c;\Theta_I)$ and the conditional generator $p(x|z,c;\Lambda)$ as a conditional variational auto-encoder (cVAE). The learning process can be reframed as cooperative learning between the cVAE $\{p(x|z,c;\Lambda), \pi(z|x,c;\Theta_I)\}$ and the conditional EBM $p(x|z,c;\Theta_E)$.

It becomes a chasing game where the cVAE endeavors to approach the conditional EBM towards the true data distribution. The loss of the EBM is the same as Eq.~(\ref{app:coop_objective}) above. The cVAE learns from the EBM's synthesized examples $\{(\tilde{x}_i, c_i),i=1,...,n\}$, that is, 
\begin{equation}
    \min_{\Lambda,\Theta}  L_{VAE}(\Lambda, \Theta_
I) = \frac{1}{\tilde{n}} \sum_{i=1}^{\tilde{n}}[- \log p(\tilde{x}_i|c_i;\Lambda) + KL(\pi(z|\tilde{x}_i,c_i;\Theta_I)||p(z|\tilde{x}_i,c_i;\Lambda))], 
\end{equation}
which is equivalent to
\begin{equation}
\min_{\Lambda,\Theta}  L_{VAE}(\Lambda, \Theta_
I) = \frac{1}{\tilde{n}} \sum_{i=1}^{\tilde{n}}[ KL(\pi(z|\tilde{x}_i,c_i;\Theta_I)||p(z)) - E_{\pi(z|\tilde{x}_i,c_i;\Theta_I)}[\log p(\tilde{x}_i|z,c_i;\Lambda)].
\end{equation}
The left term represents the KL divergence between two Gaussian distributions, which is tractable; meanwhile, the second term denotes a reconstruction loss. 

Different from the original cVAE, which learns from the fixed data distribution, our cVAE learns from the dynamic conditional EBM and simultaneously acts as a rapid initializer for the EBM's MCMC sampling. Hence, we term this approach "Variational MCMC Teaching."

\subsection{Conditional generation, Hash head, and discriminative head} 

With the conditional generator, we can randomly generate contrastive image pairs, utilizing them to construct a triplet-ranking loss for training the hash model (i.e., hash head). Additionally, the discriminative loss, leveraging label information, acts as regularization in training the descriptor.

\section{Mode Collapse: CoopHash v.s. HashGAN}~\label{sec:mode_collapse}

We will first start from the definition of the KL divergence. Suppose we have two probability densities $q$ and $p$, the KL divergence between $q$ and $p$ is defined as $KL(q|p)=E_q[\log \frac{q}{p}]$, where $E_q$ denotes the expecation under distribution $q$.

The CoopHash is a generative model based on energy-based cooperative learning, which is based on maximum likelihood estimation (MLE).

Let us use $q_{data}(x)$ to denote an unknown data distribution. Let $p_{\theta}$ be the model we need to train. For a large number of training examples, maximizing the likelihood is equivalent to minimizing the KL divergence between the data distribution and the model, i.e., $KL(q_{data}(x)|p_{\theta}(x)).$. $KL(q_{data}(x)|p_{\theta}(x))$ is called forward KL. In forward KL, we have $KL(q_{data}(x)|p_{\theta}(x))=E_{q_{data}(x)}[\log \frac{q_{data}(x)}{p_{\theta}(x)}]=\int q_{data}(x)[\log \frac{q_{data}(x)}{p_{\theta}(x)}]dx$, so the difference between $q_{data}$ and $p$ is weighted by $q_{data}$.

Consider $q_{data}(x)=0$ for a particular $x$. What does that mean? As $q_{data}(x)$ is the weight, then it doesn’t really matter what’s the value of the other term $\log \frac{q_{data}(x)}{p_{\theta}(x)}$, which is the difference between $q_{data}$ and $p_{\theta}$. In other words, if $q_{data}=0$, there is no consequence at all to have very big difference between $q_{data}$ and $p_{\theta}$ at that point. However, if $q_{data}>0$, then the term $\log \frac{q_{data}(x)}{p_{\theta}(x)}$ will contribute to the overall KL Divergence. If our objective is to minimize KL divergence, during the optimization, the difference between $q_{data}$ and $p_{\theta}(x)$ will be minimized at those locations $x$ where $q_{data}(x)>0$.

Now let us think about an example. Suppose $q_{data}$ is a two-mode data distribution, and $p_{\theta}$ is a single-mode model due to limited capacity. In this case, we know that $p$ is impossible to perfectly cover $q_{data}$. We want to point out that minimizing $KL(q_{data}(x)|p_{\theta}(x))$ will lead to that $p_{\theta}$ spreads out as much as possible to cover all $p_{data}(x)>0$ (i.e., modes). $p$ doesn't want to miss any mode of $q_{data}$. That is what we call "mode coverage". So in general, likelihood-based generative model will not drop modes due to the usage of forward KL.

As to GAN-based method, we minimize the generator $G$ along with a discriminator $D$ by an adversarial, zero-sum game. Specifically, let $G(z)=g_{\alpha}(z)$ be a generator. To be specific, adversarial learning's objective is $\min_G \min_D V(D,G)=E_{q_{data}}[\log D(x)]+E_{z \sim p(z)}[\log (1-D(G(z)))]$. $V(D,G)$ is the log-likelihood for $D$, i.e., the log-probability of the real and faked examples. However, $V(D,G)$ is not a very convincing objective for $G$. In practice, the training of $G$ is usually modified into maximizing $E_{z \sim p(z)}[\log D(G(z))]$ to avoid the vanishing gradient problem.

Let $p_{\alpha}$ be the distribution of the generator $G(z)$ in GAN. Assuming a perfect discriminator $D$. Then, according to Bayes’ theorem, $D(x) = q_{data}(x) / (q_{data}(x)+p_{\alpha}(x))$ (assuming equal numbers of real and faked examples). Then, $\alpha$ minimizes the Jensen-Shannon (JS) divergence $JS(q_{data}|p_{\alpha})=KL(p_{\alpha}| p_{mix}) + KL(q_{data}(x)|p_{mix})$, where $p_{mix}=(q_{data}+p_{\alpha})/2$. In JS divergence, the model $p_{\alpha}$ appears on the left-hand side of KL divergence, which is a reverse KL, i.e., this term $KL(p_{\alpha}|p_{mix})$.

In reverse KL, $KL(p_{\alpha}(x)|p_{mix}(x))=E_{p_{\alpha}(x)}[\log \frac{p_{\alpha}(x)}{p_{mix}(x)}]=\int p_{\alpha}(x)[\log \frac{p_{\alpha}(x)}{p_{mix}(x)}]dx$. The divergence is weighted by $p_{\alpha}$. That is, for Reverse KL, it is better to fit just some portion of the target (here is $p_{mix}$) as long as that approximate is good. Consequently, Reverse KL will try to avoid spreading the approximate. Now, there would be some portion of $p_{mix}$ that will not be covered by $p_{\alpha}$. That is the mode drop. Therefore, GAN-based training encourages the generator to fit some major modes of while ignoring others.

Due to the above difference between forward KL and reverse KL, the behaviors of HashGAN (GAN-based hashing method) and our CoopHash (using MLE) are essentially different. HashGAN drops mode but focuses on major modes; while CoopHash based on MLE seeks to cover all modes, thus avoiding mode collapse.

\section{Label-Image Mutual Information Maximization}~\label{sec:coophash_mi}

Let the distribution that approximates $p(c|x)$  as $q(c|x)$. By using Lemma A.1 in~\citep{chen2016infogan}, we can derive the following relationship:
\begin{align*}
     L_I(g,q) 
    & = E_{c\sim p(c), x \sim p_g(c,z)} [\log q(c|x)] + H(c) \\ \nonumber
    & = E_{x\sim p_g(c,z)}[E_{c'\sim p(c|x)}[\log q(c'|x)]] + H(c) \\ \nonumber
    & \le I(c; g(c,z)) \nonumber
\end{align*}
where $L_I(g,q)$ is the variational lower bound of $I(c; g(c, z; \theta_G))$, and $q(c|x)$ is the softmax cross-entropy objective in Equation (10). Therefore, optimizing the classification objective is equivalent to maximizing the variational lower bound of the mutual information.

\section{CoopHash Training Algorithm}\label{sec:training_algo}

\begin{algorithm}[h]
\caption{CoopHash learning.}
\label{code:coophash}
\begin{algorithmic}[1]
\Require
(1) training images with labels $\{(x_i,c_i), i=1,...,n\}$, (2) numbers of Langevin steps $l$, (3) number of learning iterations $T$, (4) learning rates $\gamma_D$ and $\gamma_G$.
\Ensure
(1) learning parameters of the generator $\Lambda$, (2) learning parameters of the multipurpose descriptor $\Theta=(\Theta_E, \Theta_I, \Theta_H, \Theta_C)$.
\item[]
\State $t\leftarrow 0$, initialize $\Theta$ and $\Lambda$.
\Repeat 
\State {\bf Generate synthetic contrastive image pair by the generator}: For $i = 1, ..., n$, sample a dissimilar label $c_i^-$ such that $c_i^- \ne c_i$, sample $z_i \sim \mathcal{N}(0, I_d)$, and then generate a similar image $\hat{x}_i^+ = g(c_i, z_i; \Lambda)$ and a dissimilar image $\hat{x}_i^-~= g(c_i^-, z_i; \Lambda)$ to form a triplet $(x_i, \hat{x}_i^+, \hat{x}_i^-)$. 
\State {\bf Refine synthetic images by the descriptor}: For $i = 1, ..., n$,  starting from $\hat{x}_i^+$ and $\hat{x}_i^-$, run $l$ steps of Langevin dynamics to obtain the refined images $\tilde{x}_i^+$ and $\tilde{x}_i^-$, respectively, each step following Eq.~(7). 
\State {\bf Update descriptor}: With the observed and the synthetic examples, we update $\Theta^{(t+1)} = \Theta^{(t)} - \gamma_D \mathcal{L}_{\text{Des}}'(\Theta^{(t)})$,  where $\mathcal{L}_{\text{Des}}(\Theta^{(t)})$ is defined in Eq.~(10)
\State {\bf Update generator}: With the synthetic examples, we update $\Lambda^{(t+1)} = \Lambda^{(t)} - \gamma_G \mathcal{L}_{\text{Gen}}'\Lambda^{(t)} $,  where $\mathcal{L}_{\text{Gen}}(\Lambda^{(t)})$ is defined in Eq.~(11). 
\State Let $t \leftarrow t+1$
\Until $t = T$
\end{algorithmic}
\end{algorithm}

Algorithm~\ref{code:coophash} describes the cooperative hashing network. The generator supplies initial samples for the MCMC of the solver. For each real-input condition $c_i$, we first sample the label $c_i^-$ for dissimilar sample $\hat{x}_i^-$ and the latent $z_i \sim N(0, I_d)$. For single-label datasets (i.e. MNIST, CIFAR-10, SVHN), $c_i^-$ is sampled as an out-of-category label w.r.t $c_i$ (i.e. $c_i^- \ne c_i$). For multi-label datasets (i.e. NUS-WIDE and COCO), $c_i^-$ is sampled as a multi-label 0-1 vector in such a way that $dot(c_i,c_i^-)=0$ (i.e. they have disjoint label sets). The initial similar and dissimilar samples can be generated as $\hat{x}_i^+=g(c_i, z_i; \Lambda)$ and $\hat{x}_i^-=g(c_i^-, z_i; \Lambda)$, respectively. If the current initializers are close to the current solver, then the generated $\{(\hat{x}_i^+,\hat{x}_i^-), i = 1,...,n\}$ should be good initializations for the solver to sample from $p(x|c_i, z_i;\Lambda)$, i.e., starting from the initial solutions $\{(\hat{x}_i^+,\hat{x}_i^-), i = 1, ..., n\}$, we run Langevin dynamics for $l$ steps to get the refined solutions $\{(\tilde{x}_i^+,\tilde{x}_i^-), i = 1, ..., n\}$.

\noindent \textbf{Runtime Complexity of CoopHash.} As seen in Algorithm~\ref{code:coophash}, with short-run MCMC, CoopHash's \textit{training complexity} is similar to that of GAN-based methods (e.g., HashGAN). Its \textit{inference complexity}, on the other hand, is similar to all the baselines since it only involves the hash function. 

\subsubsection{Multi-representation learning} 
Multiple representations are learned at the same time in the descriptor, while exploiting commonalities in the shared base network $f_{0}$ and distinction in different heads across representations. Even though our framework only need to learn a binary representation for hashing purpose, other auxiliary representations can help improve the main representation by leveraging the representation-specific information contained in the training signals of the related representational models. Thus, our strategy can result in improved learning efficiency and representation capacity, when compared to training the representational models separately.

\subsection{CoopHash's design improves Hash Function Learning}

The whole learning framework is based on the foundation of the conditional CoopNet, where an conditional EBM (bottom-up structure) and a conditional generator (top-down structure) are jointly trained via MCMC teaching. The resulting model is a probabilistic model $p(x|c)$, where $x$ is an image and $c$ is label. Our paper starts from the above foundation and adds the following design for hashing:
\begin{itemize}
    \item[(i)] {Adapt EBM for Hashing}: We add a hash head so that the model can be used for hashing. 
    \item[(ii)] {Generalize Hashing with EBM}: We change the generator to contrastive pair generator so that it can produce positive/negative to train the hash head; \textit{without the generator, we showed in the ablation study that the performance significantly decreases}.
    \item[(iii)] {Learn discriminative hash function}:  We add a discriminative head to make full use of the labels to regularize the training of the bottom-up network.
    \item[(iv)] {Ensure training efficiency of EBM}: We add inference head to amortize the MCMC-based inference in the original cooperative learning to accelerate the training. This is important to ensure that the proposed model can be used in a practical retrieval setting. Furthermore, this head also allows us to learn a hashing model that is robust against missing data.
    \item[(v)] We design all these heads to share a bottom-up network for memory efficiency.
\end{itemize}

To summarize, (i), (ii), (iii), and (iv) are necessary designs to turn the CoopNet into a powerful hashing model with better generalization, better out-of-distribution capacity, and better robustness against missing data; (iv) and (v) are efficiency designs; (v) is a design for better backbone . All of them are related and useful designs for our supervised hashing framework built on generative cooperative learning. 

Note that, the proposed VAE design cannot be trained independently without other CoopHash's components; otherwise, the above properties of CoopHash are not optimal. The decoder (i.e., the generator in CoopHash) of this VAE is trained in the cooperative framework, and requires access to the MCMC samples generated by the EBM. That is, the loss of the generator relies on the EBM.

\noindent \textbf{Comparison to GAN-based methods.} As both HashGAN and CoopHash rely on generated contrastive samples, the retrieval performance of the trained hash functions will degrade when these samples are not diverse (e.g., exhibiting mode collapse). In HashGAN, this problem is overcome by the 2-stage training; specifically, HashGAN-1 (1-stage training) significantly degrades (reported in the main paper), indicating severe mode collapse; we also observe in our experiments that HashGAN-1's samples also have higher FID than HashGAN's.

\section{Related Hashing Methods}\label{sec:related_hashing_methods}

Learning to hash, and especially image hashing, has been heavily investigated in both theory and practice. With the advance of deep neural networks, a plethora of image hashing methods have been proposed. These methods combine the principles of classical hashing techniques (i.e. to learn the balanced hash function) and the power of deep representation learning from the structured, image data. DH~\citep{erin2015deep} uses a deep network to learn a non-linear transformation from the image to the discrete hash codes, which is constrained to have low-quantization error, as well as balanced and independent hash bits. Essentially, DH can be seen as a non-linear extension of classical, linear methods such as ITQ~\citep{gong2013iterative} and Spectral Hashing~\citep{weiss2009spectral}. Similarly, DQN~\citep{cao2016bdeep} employs a more complex representation learning, AlexNet~\citep{krizhevsky2012imagenet}, while simultaneously controlling the quantization error of the hash codes. DSDH~\citet{li2017deep}, discussed in the main paper, also build the hash codes from feature representation extracted from a deep network but leverage the fact that the learned hash codes are ideal for classification. While this is a similar approach to our classification loss, the interpretation of our classification loss is different. Specifically, the classification loss not only models the relationship between the hash codes and the labels, but also can be shown as maximizing the mutual information between the image and its label. R2SDH~\citep{gui2018r2sdh} is related to DSDH and replaces the least square objective of DSDH by a cross-entropy term in order to achieve better robustness. While these supervised hashing methods achieve better performance, as discussed in the main paper, the availability of adequate labeled data could hinder their generalization performance in practical deployment. While we have compared CoopHash to several representative and recent methods in the hashing literature, as well as the state-of-the-art, we provide additional experiments in the next section and show the limited performance of these methods compared to newer methods and CoopHash.

Some other works are orthogonal to ours. ADSH~\citep{jiang2018asymmetric} explore a different setting where there is a large-scale constraint on the database data, which discourages learning of a symmetric hash function; i.e. the hash function is used to compute the hash codes for both query and database points before retrieval. In this asymmetric framework, ADSH learns the hash function only for the query points. However, in some applications, especially those in new domains, such query points cannot be collected before the similarity search system is deployed; thus, learning the hash function only from the database points, the assumption made in this paper, is still a more general scenario. 

\section{Detailed Experimental Setup}~\label{sec:experiment_details}

\subsection{Evaluation Baselines}
The current baselines include an extensive list of the latest and published works that are relevant to ours on the topic of learning better-generalization hash functions with limited labeled data. HashGAN~\citep{cao2018hashgan} and DSHGAN~\citep{qiu2017deep} are the latest GAN-based works focusing on the generalization problem. We also extensively compare CoopHash to other latest and major supervised hashing methods, e.g., CSQ, DBDH, GreedyHash, HashNet, DCH, DSDH, etc. (a total of 14 baselines) that cover diverse sets of techniques in supervised hashing. We also believe that our paper provides a more comprehensive comparison to the related works than most papers in the supervised hashing domain.   
 
\begin{table}[!h]
    \footnotesize
    \setlength{\tabcolsep}{5pt}
    \centering
    \caption{The network architectures used for CoopHash. For all the datasets, we adopt the same architecture.}
    \label{tab:network_architectures}
    \begin{tabular}{|l|l|l|l|}
    \hline
    Layers     & In-Out Size & Stride & Padding \\
    \hline
    \multicolumn{4}{|l|}{\textbf{Generator}: ngf=32, n\_classes} \\
    \hline 
    Input: $z$ & 1x1x200 & - & -\\
    Input: $c$ & 1x1x(n\_classes)  & - & -\\
    4x4 convT, LReLU, BN & 4x4x(ngf x 8) & 1 & 0 \\
    5x5 convT, LReLU, BN & 8x8x(ngf x 4) & 2 & 2 \\
    5x5 convT, LReLU, BN & 16x16x(ngf x 2) & 2 & 2 \\
    5x5 convT, LReLU & 32x32x3 & 2 & 2 \\
    \hline
    \multicolumn{4}{|l|}{\textbf{Descriptor}: ndf=32} \\
    \hline 
    Input: $z$ & 1x1x200 & - & \\
    5x5 conv, LReLU  & 15x15x(ndf x 2) & 2 & 1 \\
    3x3 convT, LReLU & 8x8x(ndf x 4) & 2 & 1 \\
    3x3 convT, LReLU & 8x8x(ndf x 8) & 1 & 1 \\
    \hline
    \multicolumn{4}{|l|}{\textbf{EBM Head}} \\
    \hline 
    256 Linear, LReLU & 256 & - & - \\
    1 & 256 & - & - \\
    \hline
    \multicolumn{4}{|l|}{\textbf{Hash Head}: n\_classes, h\_dim} \\
    \hline 
    256 Linear, LReLU & 256 & - & - \\
    h\_dim & h\_dim & - & - \\
    n\_classes &  n\_classes & - & - \\
    \hline
    \end{tabular}

\end{table}
\begin{figure*}[!tpbh]
\centering
\mbox{\hspace{-0.3in}
 \renewcommand{\thesubfigure}{a}
 \subfloat[HashNet \label{subfig:result_hash_tsne_dsh}]{%
  \includegraphics[width=0.24 \textwidth]{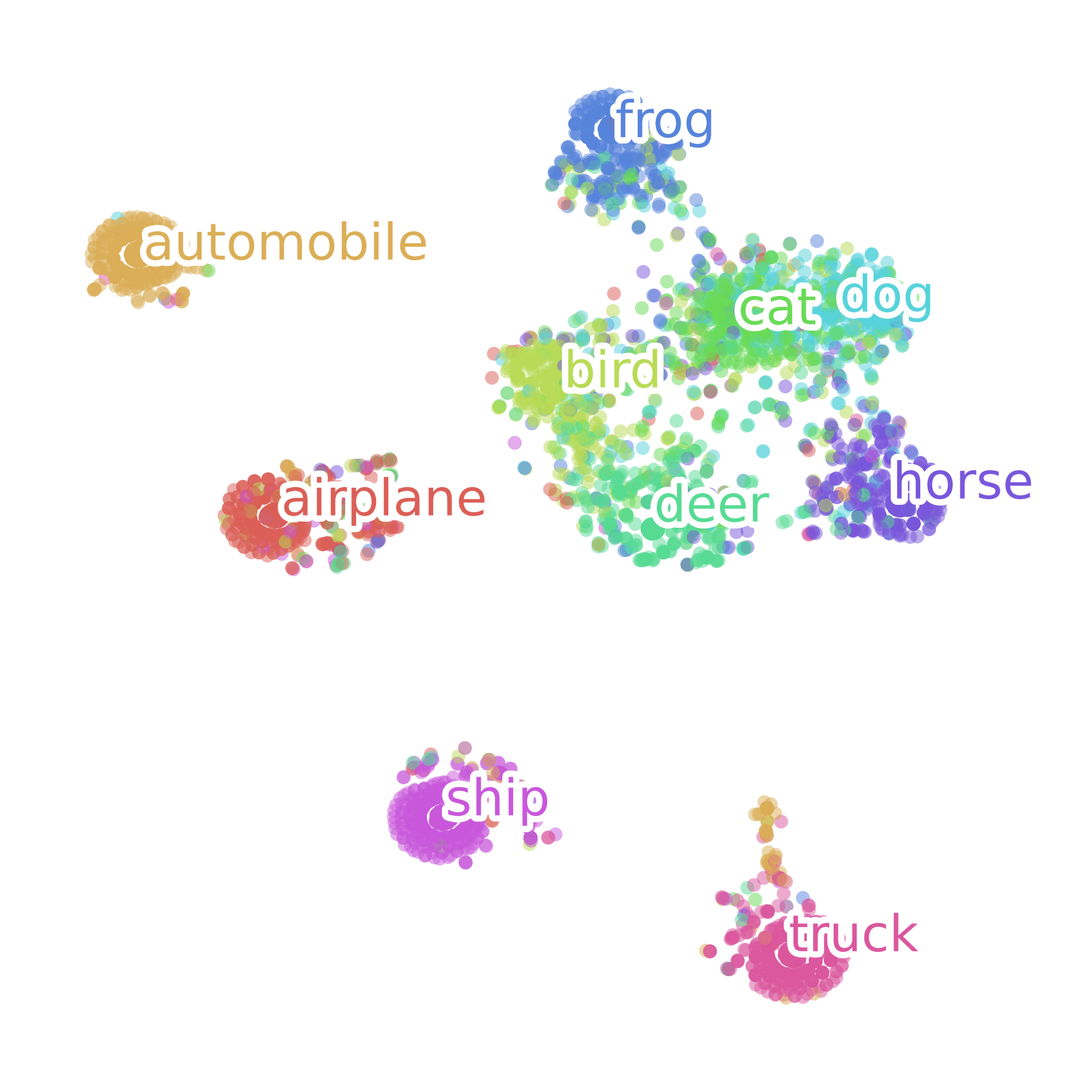}
 }
  \hfill
 \renewcommand{\thesubfigure}{b}
 \subfloat[CSQ \label{subfig:result_hash_tsne_csq}]{%
  \includegraphics[width=0.24 \textwidth]{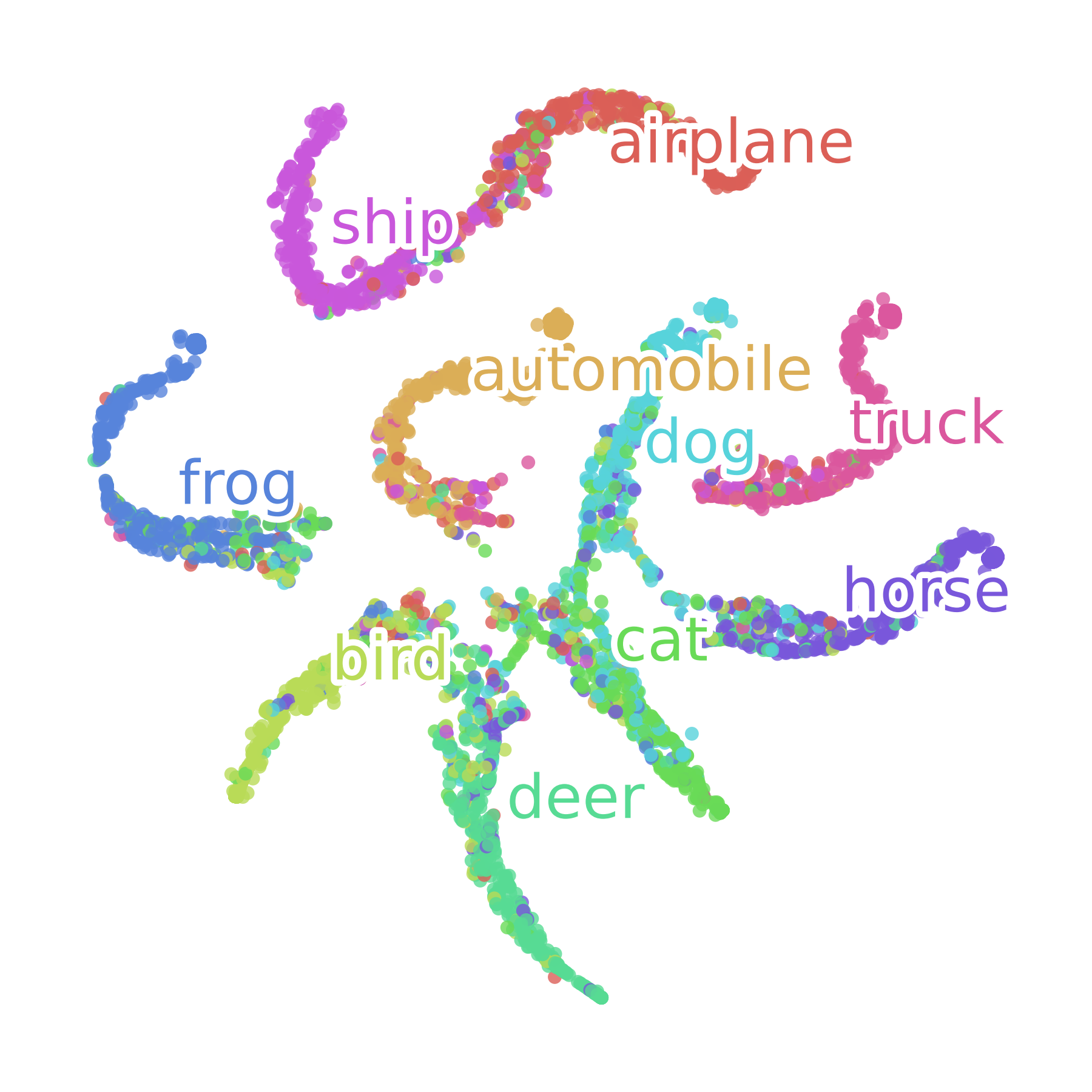}
 }
 \hfill
  \renewcommand{\thesubfigure}{c}
 \subfloat[HashGAN \label{subfig:result_hash_tsne_hashgan}]{%
  \includegraphics[width=0.24 \textwidth]{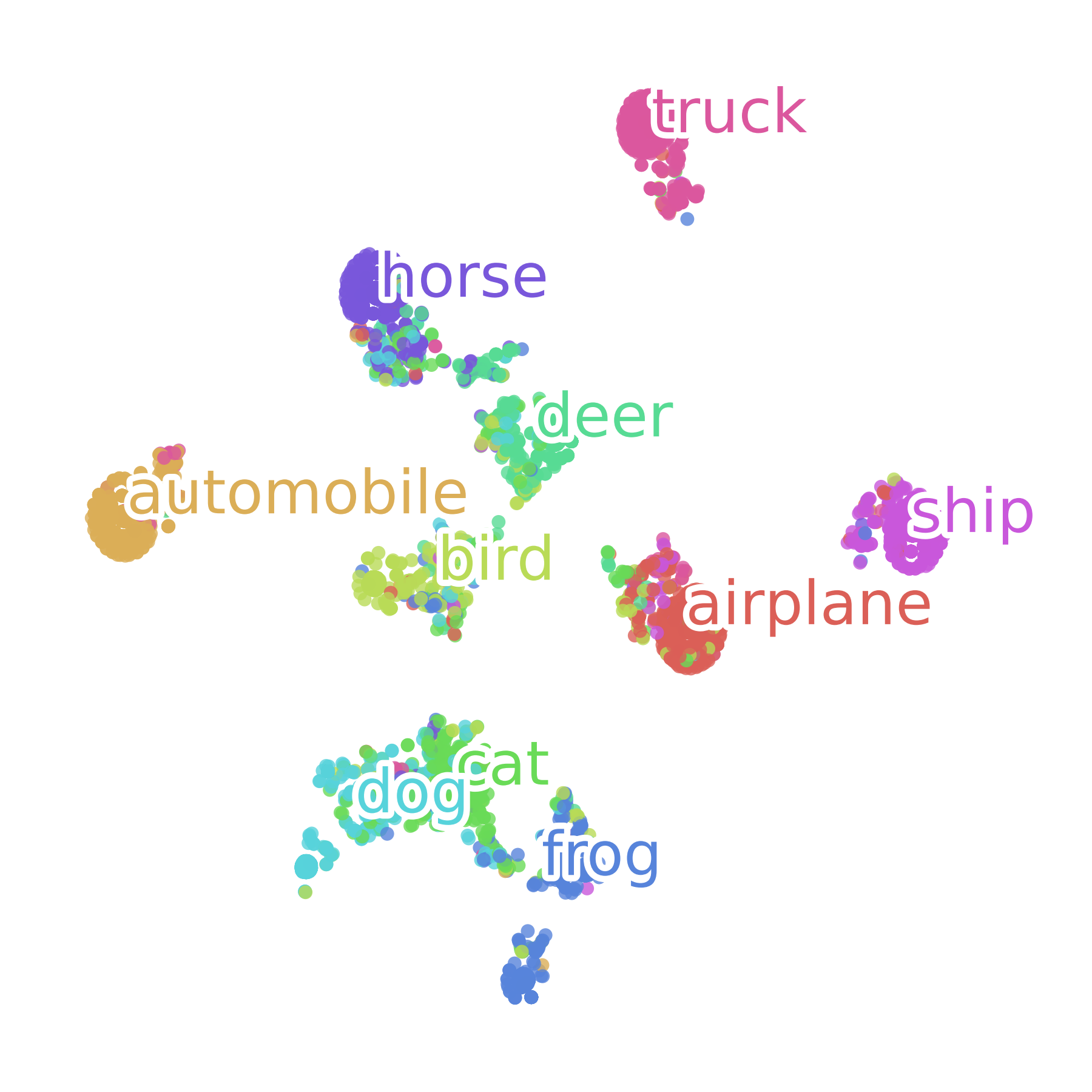}
 }
 \hfill
  \renewcommand{\thesubfigure}{d}
 \subfloat[CoopHash \label{subfig:result_hash_tsn_coophash}]{%
  \includegraphics[width=0.24 \textwidth]{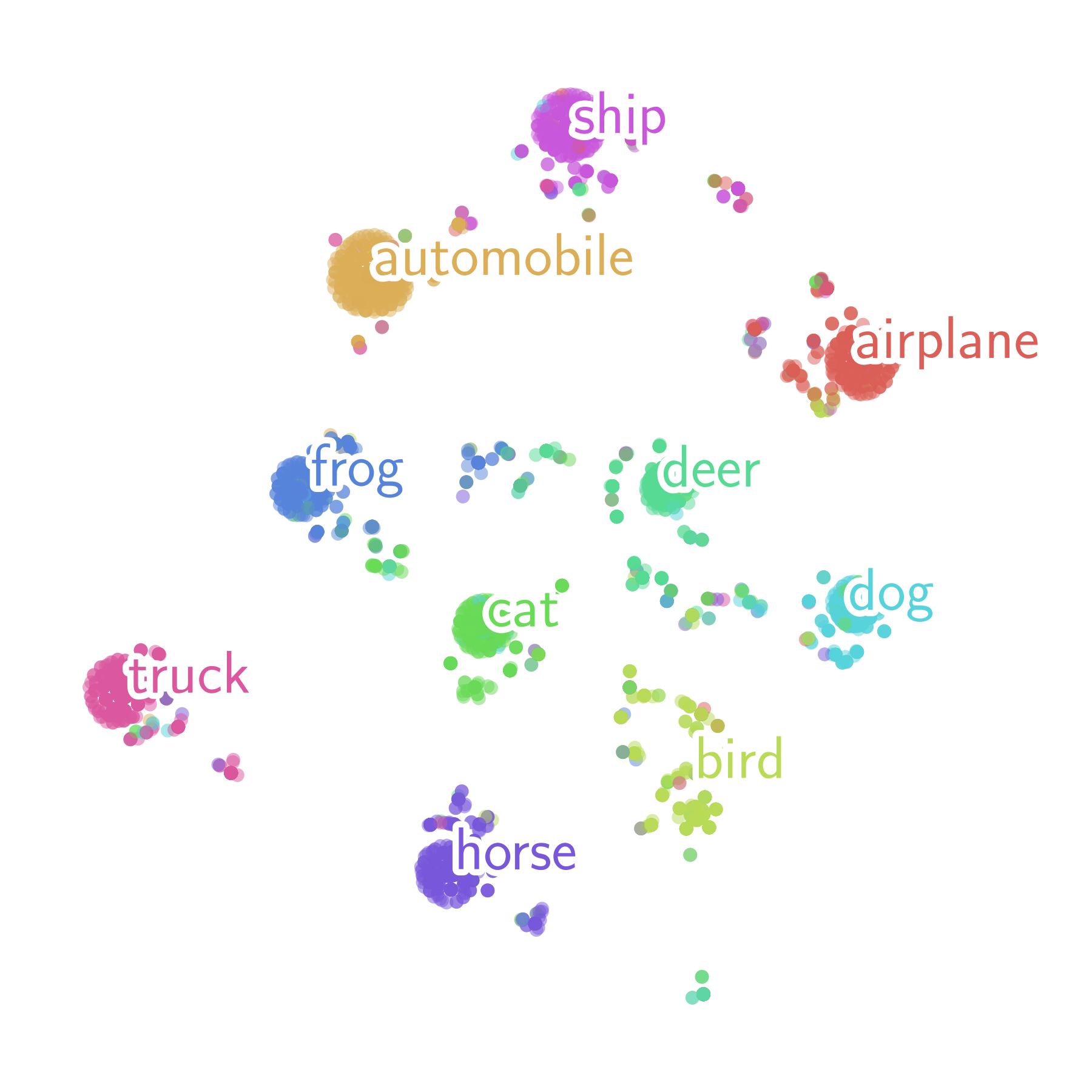}
 } 
}
\caption{The t-SNE visualizations of the quantized 32-bit hash codes learned by HashNet, HashGAN and CoopHash on the CIFAR-10 dataset.}
\label{fig:result_hash_tsne}
\end{figure*}
\begin{table*}[h]
    \footnotesize
    \centering
    \caption{Additional mAP results.}
    \begin{tabular}{lcccccc}
        \toprule
         Method & \multicolumn{3}{c}{CIFAR10} & \multicolumn{3}{c}{NUS-WIDE} \\
          & 16 bits & 32 bits & 64 bits & 16 bits & 32 bits & 64 bits \\ 
         \hline
         DQN \citep{cao2016deep} & 0.55 & 0.56 & 0.58 & 0.62 & 0.63 & 0.65 \\
         DH \citep{erin2015deep} & 0.18 & 0.20 & 0.22 & 0.45 & 0.47 & 0.50 \\
         DSDH \citep{li2017deep} & 0.70 & 0.72 & 0.74 & 0.71 & 0.72 & 0.74\\ 
         R2SDH \citep{gui2018r2sdh} & 0.42 & 0.44 & 0.52 & 0.46 & 0.49 & 0.50 \\ 
         \hline
         CoopHash & \textbf{0.73} & \textbf{0.77} & \textbf{0.79} & \textbf{0.76} & \textbf{0.77} & \textbf{0.80} \\
         \bottomrule
    \end{tabular}
    \label{tab:result_additional_mAP}
\end{table*}

\subsection{Dataset Details}
In this section, we provide the detailed description of the datasets used to evaluate the methods in our paper. Similar to~\citep{cao2018hashgan}, we construct the training dataset with limited number of samples to ensure that our experiments resembles the real-world environment where there are limited number of labeled samples. 

\noindent \textbf{NUS-WIDE} dataset contains 269,648 images, each of which belongs to at least one of the 81 concepts. We randomly select 5,000 images for the query set, with the remaining images used as the retrieval set. 10,000 images randomly selected in the retrieval set are used for training.

\noindent \textbf{COCO} dataset contains 123,287 images, labeled with at least 1 out of 80 semantic concepts. Similarly, the query set is randomly constructed with 5,000 images, with the remaining images used as the retrieval set (10,000 randomly selected images in this set are used for training).

\noindent \textbf{CIFAR-10} contains 60,000 images organized into 10 semantic classes, out of which 1,000 images are randomly selected as the query set, and the remaining images used as the retrieval set. Similarly, 5,000 images randomly selected from the retrieval set are used for training. 

\noindent \textbf{MNIST} contains 70,000 images organized into 10 digit classes, out of which  1,000 images are randomly selected as the query set, and the remaining images used as the retrieval set. The training set contains 10,000 images randomly selected from the retrieval set.

\noindent \textbf{SVHN} contains 600,000 images organized into 10 digit classes, out of which 5,000 images are randomly selected as the query set, and the remaining images used as the retrieval set. The training set contains 10,000 images randomly selected from the retrieval set.

\subsection{Implementation Details}\label{sec:implementation_details}

Following the evaluation approaches in~\citep{cao2018hashgan,cao2017hashnet}, for the shallow-hashing methods, we extract the image features (4096-dimensional vectors) from DeCAF7~\citep{donahue2014decaf}. Similarly, for the deep-hashing methods, we use AlexNet~\citep{krizhevsky2017imagenet} as the backbone. For HashGAN, we follow the original paper and employ a four-layer ResNet architecture~\citep{he2016deep} for the discriminator and generator in HashGAN. 

For the proposed method, CoopHash, the architectures of the different networks are in Table~\ref{tab:network_architectures}. To train CoopHash, we adopt Adam optimizer (beta1=0.9 and beta2=0.999) with a batch size of 64 and a learning rate of $1e-3$ for both the generation and description modules. We run 20 steps MCMC Langevin dynamics to train the EBM descriptor with a learning rate of 0.5, and standard deviation of 0.0005 for the white noise. 

All experiments are running on servers equipped with NVidia V100 or A100 GPUs.  

\section{Additional Experiments}~\label{sec:additional_experiments}


\subsection{Additional retrieval comparisons}

In this section, we provide an additional comparison to the more related methods mentioned in Section~\ref{sec:related_hashing_methods}. CoopHash consistently achieves better performance compared to these methods, as shown in Table~\ref{tab:result_additional_mAP}.

A similar superior performance of CoopHash to the results in the main paper can be observed in Table~\ref{tab:result_precision}, using the P@1000 metric. This makes CoopHash also desirable for practical, precision-oriented retrieval systems. 

\begin{table}[!t]
\setlength{\tabcolsep}{3pt}
\footnotesize
    \centering
    \caption{P@1000 on the three image datasets.}
    \begin{tabular}{l|cc|cc|cc}
        \specialrule{.1em}{.1em}{.1em} 
        
        \multirow{2}{*}{Method} & \multicolumn{2}{|c}{NUS-WIDE} & \multicolumn{2}{|c}{CIFAR-10} & \multicolumn{2}{|c}{COCO} \\
        \cline{2-7}
        & 32  & 64  & 32  & 64  & 32  & 64   \\ 
        
        \hline
            
        ITQ & 0.572 & 0.590 & 0.271 & 0.305 & 0.518 & 0.545  \\ 
        
        BRE & 0.603 & 0.627 & 0.445 & 0.471 & 0.535 & 0.559  \\ 
        
        KSH & 0.656 & 0.667 & 0.612 & 0.641 & 0.540 & 0.558  \\ 
        
        SDH & 0.702 & 0.712 & 0.671 & 0.651 & 0.695 & 0.710  \\ 
        
        CNNH & 0.651 & 0.672 & 0.545 & 0.578 & 0.690 & 0.718  \\ 
        
        DNNH & 0.689 & 0.707 & 0.678 & 0.691 & 0.701 & 0.728  \\ 
        
        DHN & 0.713 & 0.726 & 0.701 & 0.725 & 0.731 & 0.750  \\  
        
        DSDH & 0.728 & 0.752 & 0.710 & 0.729 & 0.735 & 0.754  \\  
        
        HashNet & 0.729 & 0.741 & 0.721 & 0.741 & 0.746 & 0.753  \\
        
        DCH & 0.723 & 0.731 & 0.691 & 0.712 & 0.551 & 0.584  \\
        
        GreedyHash & 0.719 & 0.728 & 0.712 & 0.726 & 0.539 & 0.574  \\
        
        DBDH & 0.739 & 0.745 & 0.597 & 0.602 & 0.582 & 0.591  \\
        
        CSQ & 0.741 & 0.750 & 0.735 & 0.741 & 0.742 & 0.749  \\
        \hline

        DSHGAN & 0.763 & 0.780 & 0.759 & 0.781 & 0.765 & 0.785  \\ 
        
        HashGAN & 0.759 & 0.772 & 0.751 & 0.762 & 0.768 & 0.783  \\ 
        
        CoopHash & \textbf{0.791} & \textbf{0.810} & \textbf{0.803} & \textbf{0.829} & \textbf{0.801} & \textbf{0.820}  \\  
        \specialrule{.1em}{.1em}{.1em} 
    \end{tabular}
    \label{tab:result_precision}
\end{table}

\subsection{Visualization of the hash Codes}

Figure~\ref{fig:result_hash_tsne} shows the 2-dimensional t-SNE embedding of the learned hash codes of the proposed approach, HashNet, CSQ, and HashGAN. We observe that the learned hash codes of HashNet and CSQ exhibit a more convoluted structure than the learned hash codes of  both HashGAN and CoopHash. CoopHash exhibits a clearer discriminative structure in the hash codes, compared to other methods. This demonstrates the effectiveness of the proposed method, which leads to a better retrieval performance.

\subsection{Effect of Number of Langevin Steps}
\begin{figure}[h]
    \centering
    \includegraphics[width=0.45 \textwidth]{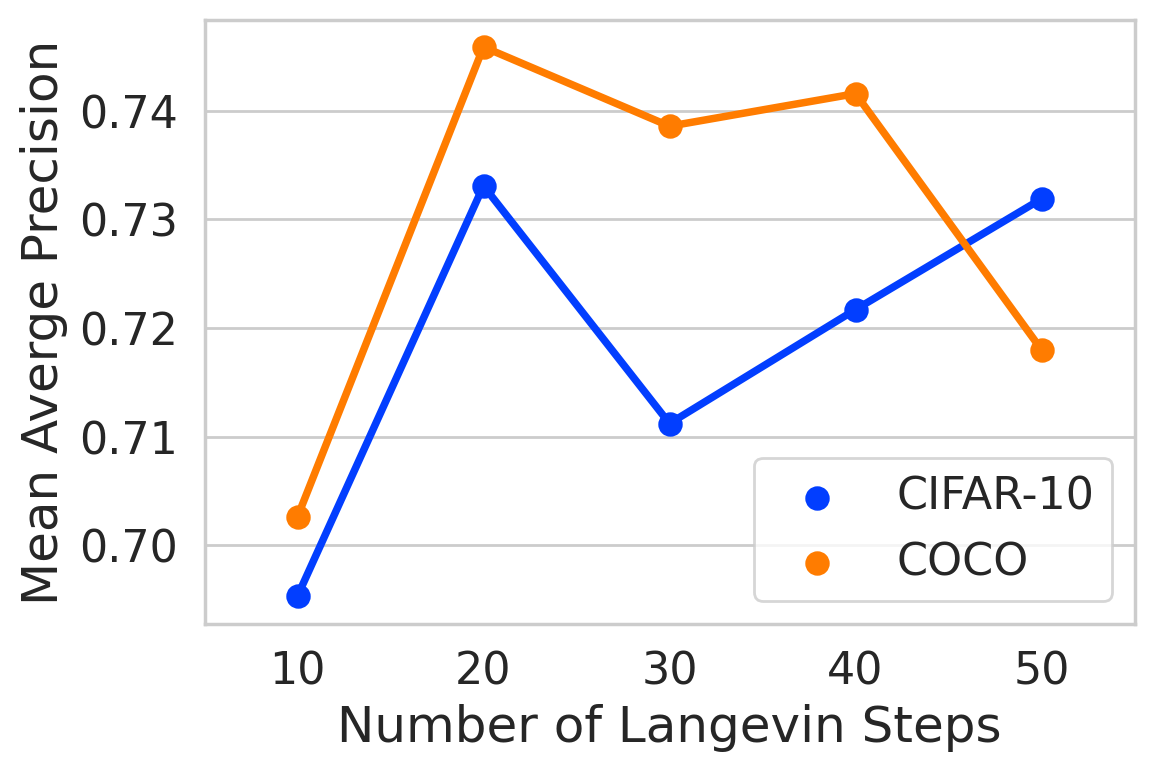}
    \caption{Retrieval Performance when varying the Number of Langevin Steps}
    \label{fig:result_ablation_langevin_num_steps}
\end{figure}

In this section, we present the results to determine the influence of the number of MCMC steps to the retrieval performance. Figure~\ref{fig:result_ablation_langevin_num_steps} shows the mAP results for different numbers of steps. We can observe that when the number of MCMC steps reach a certain values (20 or more), the retrieval performance reaches optimal values and increasing the number of steps does not significantly change the performance. The small number of steps to reach a good performance shows that we can efficiently train CoopHash with short-run MCMC.

\subsection{Effect of Generator's Input Latent's Dimensions}
\begin{figure}[h]
    \centering
    \includegraphics[width=0.45 \textwidth]{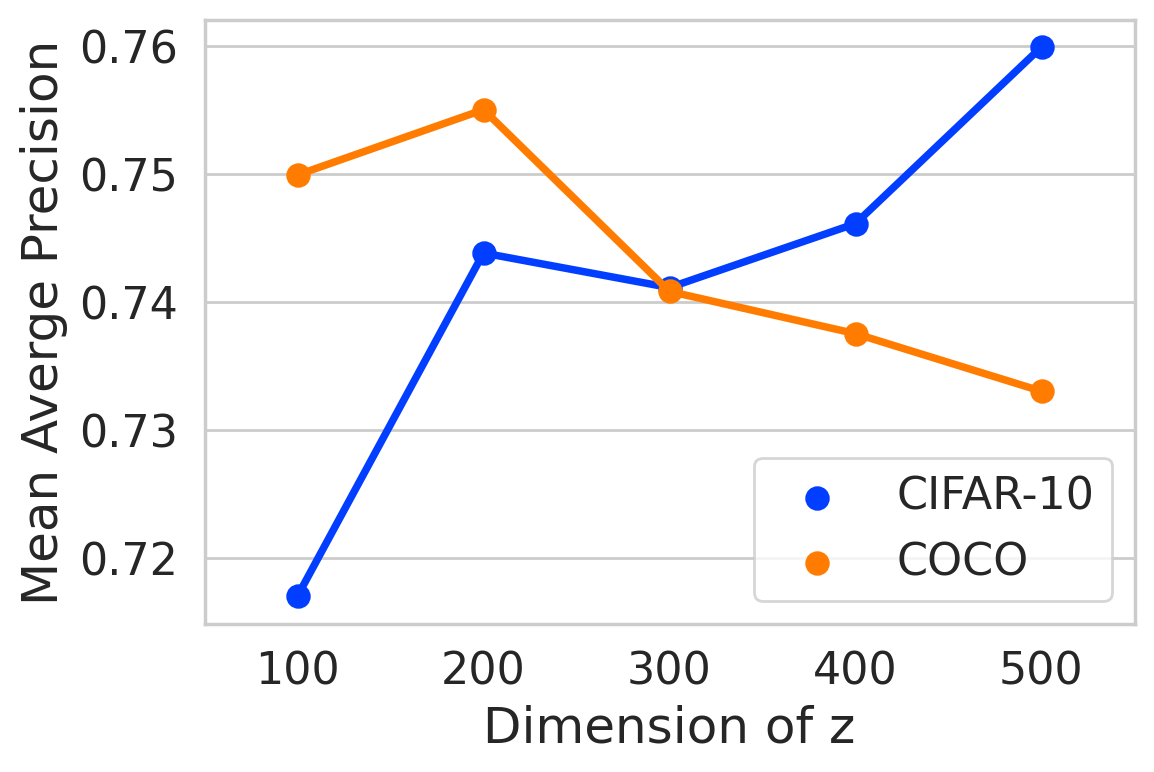}
    \caption{Retrieval Performance when varying the size of the Latent Input to the Generator}
    \label{fig:result_ablation_latent_dim}
\end{figure}
\begin{figure*}[!h]
 \centering
 \renewcommand{\thesubfigure}{a}
 \subfloat[Real Images \label{subfig-1:sqrt_errors}]{%
  \includegraphics[width=0.45 \textwidth]{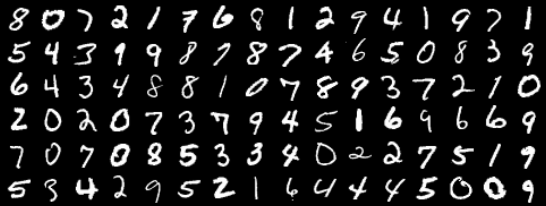}
 }
 \hfill
 \renewcommand{\thesubfigure}{b}
 \subfloat[Generated Images \label{subfig-2:sqrt_gradients}]{%
  \includegraphics[width=0.45 \textwidth]{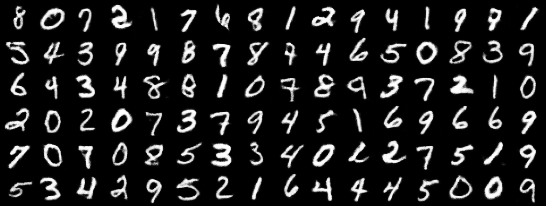}
 }
 \caption{Generated MNIST Digits, conditioned on the label. The images in the same position are conditioned on the same label.}
 \label{fig:result_generated_images_mnist}
\end{figure*}

\begin{figure*}[!htpb]
 \centering
 \renewcommand{\thesubfigure}{a}
 \subfloat[Real Images \label{subfig-1:sqrt_errors}]{%
  \includegraphics[width=0.45 \textwidth]{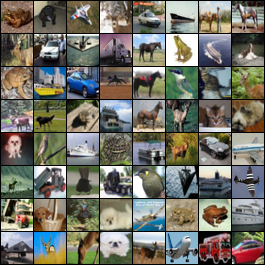}
 }
 \hfill
 \renewcommand{\thesubfigure}{b}
 \subfloat[Generated Images \label{subfig-2:sqrt_gradients}]{%
  \includegraphics[width=0.45 \textwidth]{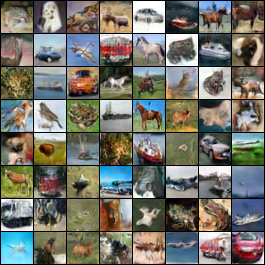}
 }
 \caption{Generated CIFAR-10 images, conditioned on the label. The images in the same position are conditioned on the same label.}
 \label{fig:result_generated_images_cifar10}
\end{figure*}
In this section, we present the results to determine the influence of size of the latent input into the generator. Figure~\ref{fig:result_ablation_latent_dim} shows the mAP results for different sizes of latent vector $z$. In general, we can observe that the latent dimension does not influence the optimal performance significantly.

\subsection{Generating Conditional Images}
In this section, we show the generated MNIST digits and CIFAR-10 by CoopHash. Note that, we can use a better generator's architecture for CoopHash, which can generate better images. In this paper, the selected resolution is 32x32.

Figures~\ref{fig:result_generated_images_mnist} and~\ref{fig:result_generated_images_cifar10} show the generated images. In general, CoopHash successfully generates images for each conditional category. While CoopHash learns a good hash function, this experiment demonstrates that CoopHash can jointly learn both objectives of good image generation and hash function learning.

\subsection{Robustness toward missing data}

In CoopHash, we estimate a joint probability of the image and its label through the energy head. This formulation also allows us to reconstruct from the corrupted samples by way of the Langevin dynamics by feeding these corrupted images as initial samples instead of samples from the generator. In addition, while the inference head assists in learning the generator function, the resulting VAE network can also be trained to perform image-recovery tasks such as image in-painting. This means that CoopHash can also handle corrupted data in both training and testing.

Specifically, we assume that both training data and testing data may contain corrupted input images. During the training of CoopHash, we train on the clean input as mentioned previously. For corrupted input, using the proposed VAE, we compute the reconstructed samples. We treat these reconstructed samples as synthesized samples which are then updated through the Langevin dynamics of the descriptor. Similar to training of clean input, we learn the parameters of the VAE network by minimizing $\mathcal{L}_{VAE}$ in Eq. (10). \textbf{Thus, there is essentially no changes in the architecture and loss functions of CoopHash}. Intuitively, this process teaches the VAE network to learn to in-paint the corrupted images. Note that, this is not possible with other methods and especially a GAN-based approach such as in HashGAN. 

In Table~\ref{tab:result_additional_corruption}, we show the robustness of our method on CIFAR10 when corrupted inputs are present during training and testing. Similar to the other experiments, we use the same number of samples for train and test, but corrupted 20\% of the data using salt-and-pepper (denoted by SnP) noise and random rectangular mask (denoted by RRM, where the rectangles randomly cover approximately 10-20\% of the images at random locations) on the images for both train and query sets. 

As observed, CoopHash is more robust than other hashing methods when the data is corrupted. This shows the advantages of the proposed multipurpose descriptor/generator hashing network.

\begin{table}[]
    \centering
    \caption{mAP results for data corruption experiments (32 bits).}
    \mbox{\hspace{-15pt}
    \begin{tabular}{llccc}
        \toprule
         Corruption & Type & HashNet & HashGAN & CoopHash \\ \hline
         \multirow{2}{*}{SnP} & Clean & 0.51 & 0.60 & 0.68  \\
          & Corrupted & 0.43 & 0.49 & 0.65  \\ \hline
         \multirow{2}{*}{RRM} & Clean & 0.47 & 0.61 & 0.65  \\
          & Corrupted & 0.39 & 0.48 & 0.60  \\
         \bottomrule
    \end{tabular}
    }
    \label{tab:result_additional_corruption}
\end{table}

\subsection{Retrieval in Unseen Setting}
\begin{table}[]
    \centering
    \caption{Retrieval in unseen setting on CIFAR10}
    \label{tab:result_retrieval_unseen}
    \begin{tabular}{lccc}
        \toprule
         Method & 16 bits & 32 bits & 64 bits \\ 
         \hline
         HashNet & 0.667 & 0.778 & 0.825  \\
         DSDH & 0.758 & 0.769 & 0.777 \\
         CSQ & 0.781 & 0.778 & 0.789 \\
         HashGAN & 0.795 & 0.802 & 0.820 \\
         CoopHash & 0.809 & 0.814 & 0.835 \\
         \bottomrule
    \end{tabular}
\end{table}

We evaluated the unseen setting, introduced in~\citep{sablayrolles2017should},  on CIFAR10 in Table~\ref{tab:result_retrieval_unseen}. Similar to~\citep{sablayrolles2017should}, we divide the dataset into two disjoint subsets: one subset is used to train the hash methods while retrieval is performed on the other subset. CoopHash performs better, which is consistent with other types of ``unseen'' setting reported in the paper, such as data shift and corruption.

\section{Limitations} \label{sec:limitations}

This paper presented a generative approach to learn a hash function. To the best of our knowledge, this paper presented the first generative energy-based approach that can generate additional data and learn powerful discrete representations in order to improve retrieval performance. Our study mainly targets the deep supervised image hashing domain where the labeled similarity data is limited.  However, since this is the first work on generative energy-based models in image hashing, we believe there is a large room for improving both the model architectures and computational performance, including both training and inference speeds. We hope our study can serve as an effective baseline for future works in this direction.

%


\end{document}